\documentclass[10pt,twocolumn]{article}
\usepackage{graphicx}
\usepackage{amsmath,amssymb,times,cite}
\usepackage{xcolor}
\usepackage{booktabs}

\newcommand{\op}[1]{\operatorname{#1}}
\newcommand{\bm}[1]{\mathbf{#1}} 

\newcommand\T{{\mathpalette\raiseT\intercal}}
\newcommand\raiseT[2]{%
\setbox0\hbox{$#1{#2}$}\raise\dp0\box0}

\title{\Large\textbf{Graph Fairing Convolutional Networks for Anomaly Detection}}
\author{Mahsa Mesgaran and A. Ben Hamza\\
Concordia Institute for Information Systems Engineering\\
Concordia University, Montreal, QC, Canada
}
\date{}

\begin{document}
\maketitle

\begin{abstract}
Graph convolution is a fundamental building block for many deep neural networks on graph-structured data. In this paper, we introduce a simple, yet very effective graph convolutional network with skip connections for semi-supervised anomaly detection. The proposed layerwise propagation rule of our model is theoretically motivated by the concept of implicit fairing in geometry processing, and comprises a graph convolution module for aggregating information from immediate node neighbors and a skip connection module for combining layer-wise neighborhood representations. This propagation rule is derived from the iterative solution of the implicit fairing equation via the Jacobi method. In addition to capturing information from distant graph nodes through skip connections between the network's layers, our approach exploits both the graph structure and node features for learning discriminative node representations. These skip connections are integrated by design in our proposed network architecture. The effectiveness of our model is demonstrated through extensive experiments on five benchmark datasets, achieving better or comparable anomaly detection results against strong baseline methods. We also demonstrate through an ablation study that skip connection helps improve the model performance.
\end{abstract}


\section{Introduction}
Anomaly detection is of paramount importance when deploying artificial intelligence systems, which often encounter unexpected or abnormal items/events that deviate significantly from the majority of data. Anomaly detection techniques are widely used in a variety of real-world applications, including, but not limited to, intrusion detection, fraud detection, healthcare, Internet of Things, Industry 4.0 and beyond, surveillance, and social networks~\cite{Chandola:09,Guansong:20,Doshi:21}. Most of these techniques often involve a training set where no anomalous instances are encountered, and the challenge is to identify suspicious items or events, even in the absence of abnormal instances. In practical settings, the output of an anomaly detection model is usually an alert that triggers every time there is a anomaly or a pattern in the data that is atypical.

Detecting anomalies is a challenging task, primarily because the anomalous instances are not known a priori and also the vast majority of observations in the training set are normal instances. Therefore, the mainstream approach in anomaly detection has been to separate the normal instances from the anomalous ones by using unsupervised learning models. One-class support vector machines (OC-SVM) are a classic example~\cite{Scholkopf:01}, which is a one-class classification model trained on data that has only one class (i.e. normal class) by learning a discriminative hyperplane boundary around the normal instances. Another commonly-used anomaly detection approach is support vector data description (SVDD)~\cite{tax2004support}, which basically finds the smallest possible hypersphere that contains all instances, allowing some instances to be excluded as anomalies. Zhang \textit{et al.}~\cite{Zhang2023Graph} presented a graph model-based multiscale feature fitting method for unsupervised anomaly detection and localization. Arias \textit{et al.}~\cite{Arias2023AIDA} introduced an unsupervised parameter-free analytic isolation and distance-based anomaly detection algorithm, which integrates both distance and isolation metrics. However, these approaches rely on hand-crafted features, are unable to appropriately handle high-dimensional data, and often suffer from computational scalability issues.

Deep learning has recently emerged as a very powerful way to hierarchically learn abstract patterns from data, and has been successfully applied to anomaly detection, showing promising results in comparison with shallow methods~\cite{wang2020deep}. Ruff \textit{et al.}~\cite{ruff2018deep} extend the shallow one-class classification SVDD approach to the deep learning setting by proposing a deep learning based SVDD framework for anomaly detection using an anomaly detection based objective. Deep SVDD is an unsupervised learning model that learns to extract the common factors of variation of the data distribution by training a neural network while minimizing the volume of a hypersphere that encloses the network representations of the data. Also, Ruff \textit{et al.}~\cite{ruff2019deep} introduce a deep semi-supervised anomaly detection (Deep SAD) approach, which is a generalization of the unsupervised Deep SVDD technique to the semi-supervised setting. Deep SAD differs from Deep SVDD in that its objective function also includes a loss term for labeled normal and anomalous instances. The more diverse the labeled anomalous instances in the training set, the better the anomaly detection performance of Deep SAD. The key difference between deep one-class models such as Deep SVVD and semi-supervised anomaly detection methods such as Deep SAD lies in the way they are trained and the amount of labeled data required. The former is an unsupervised anomaly detection method that requires only normal data during training. It is trained to learn a representation of normal data, which is then used to distinguish between normal and anomalous data. Semi-supervised anomaly detection, on the other hand, is a hybrid approach that generally uses both normal and anomalous data during training. The model is trained using both labeled and unlabeled data, where the labeled data consists of a small portion of anomalous data and a large portion of normal data. The main advantage of semi-supervised anomaly detection is that it can achieve higher accuracy than unsupervised methods, as the use of labeled data during training provides additional information, enabling the model to better distinguish between normal and anomalous data points.

Owing to the recent developments in deep semi-supervised learning on graph-structured data, there has been a surge of interest in the adoption of graph neural networks for learning latent representations of graphs~\cite{Defferrard:16,Kipf:17}. Defferrard \textit{et al.}~\cite{Defferrard:16} introduce the Chebyshev network, an efficient spectral-domain graph convolutional neural network that uses recursive Chebyshev polynomial spectral filters to avoid explicit computation of the Laplacian spectrum. These filters are localized in space, and the learned weights can be shared across different locations in a graph. An efficient variant of graph neural networks is graph convolutional networks (GCNs)~\cite{Kipf:17}, which is an upsurging semi-supervised graph-based deep learning framework that uses an efficient layer-wise propagation rule based on a first-order approximation of spectral graph convolutions. The feature vector of each graph node in GCN is updated by essentially applying a weighted sum of the features of its immediate neighboring nodes. While significant strides have been made in addressing anomaly detection on graph-structured data~\cite{akoglu2015graph}, it still remains a daunting task on graphs due to various challenges, including graph sparsity, data nonlinearity, and complex modality interactions~\cite{ding2019deep}. Ding~\textit{et al.}~\cite{ding2019deep} design a GCN-based autoencoder for anomaly detection on attributed networks by taking into account both topological structure and nodal attributes. The encoder of this unsupervised approach encodes the attribute information using the output GCN embedding, while the decoder reconstructs both the structure and attribute information using non-linearly transformed embeddings of the output GCN layer. The basic idea behind anomaly detection methods based on reconstruction errors is that the normal instances can be reconstructed with small errors, while anomalous instances are often reconstructed with large errors. More recently, Kumagai~\textit{et al.}~\cite{kumagai2020semi} have proposed two GCN-based models for semi-supervised anomaly detection. The first model uses only labeled normal instances, whereas the second one employs labeled normal and anomalous instances. Both models are trained to minimize the volume of a hypersphere that encloses the GCN-learned node embeddings of normal instances, while embedding the anomalous ones outside the hypersphere.

Inspired by the implicit fairing concept in geometry processing for triangular mesh smoothing~\cite{Desbrun:99}, we introduce a graph fairing convolutional network architecture, which we call GFCN, for deep semi-supervised anomaly detection. In addition to performing graph convolution, GFCN uses a skip connection to combine both the initial node representation and the aggregated node neighborhood representation, enabling it to memorize information across distant nodes. While most graph convolutions with skip connections are based on heuristics, GFCN is theoretically motivated by implicit fairing and derived from the Jacobi iterative method. In contrast to GCN-based methods that use a first-order approximation of spectral graph convolutions and a renormalization trick in their layer-wise propagation rules to avoid numerical instability, our GFCN model does not require any renormalization, while still maintaining the key property of convolution as a neighborhood aggregation operator. Hence, repeated application of the GFCN's layer-wise propagation rule provides a computationally efficient convolutional process, leading to numerical stability while avoiding the issue of exploding/vanishing gradients. The proposed framework achieves better anomaly detection performance, as GFCN uses a multi-layer architecture, together with skip connections, and non-linear activation functions to extract high-order information of graphs as discriminative features. Multi-layer architectures enable the model to learn hierarchical representations of the graph, where lower layers capture lower-level features and higher layers capture higher-level abstractions. Skip connections allow information to bypass intermediate layers and preserve low-level details, improving the flow of information and preventing vanishing gradients, and more importantly leading to more accurate representations, thereby yielding a more effective detection of anomalies. Moreover, GFCN inherits all benefits of GCNs, including accuracy, efficiency and ease of training.

In addition to capturing information from distant graph nodes through skip connections between layers, the proposed GFCN model is flexible and exploits both the graph structure and node features for learning discriminative node representations in an effort to detect anomalies in a semi-supervised setting. Not only does GFCN outperforms strong anomaly detection baselines, but it is also surprisingly simple, yet very effective at identifying anomalies. The main contributions of this work can be summarized as follows:
\begin{itemize}
\item We propose a novel multi-layer graph convolutional network with a skip connection for semi-supervised anomaly detection by effectively exploiting both the graph structure and attribute information.
\item We introduce a learnable skip-connection module, which helps nodes propagate through the network's layers and hence substantially improves the quality of the learned node representations.
\item We analyze the complexity of the proposed model and train it on a regularized, weighted cross-entropy loss function by leveraging unlabeled instances to improve performance.
\item We demonstrate through extensive experiments that our model can capture the anomalous behavior of graph nodes, leading to state-of-the-art performance across several benchmark datasets.
\end{itemize}	

The rest of this paper is organized as follows. In Section 2, we review important relevant work. In Section 3, we outline the background for spectral graph theory and present the problem formulation. In Section 4, we introduce a graph convolutional network architecture with skip connection for deep semi-supervised anomaly detection. In Section 5, we present experimental results to demonstrate the competitive performance of our approach on five standard benchmarks. Finally, we conclude in Section 6 and point out future work directions.

\section{Related Work}
The basic goal of anomaly detection is to identify abnormal instances, which do not conform to the expected pattern of other instances in a dataset. To achieve this goal, various anomaly detection techniques have been proposed, which can distinguish between normal and anomalous instances~\cite{Chandola:09}. Most mainstream approaches are one-class classification models~\cite{Scholkopf:01,tax2004support} or graph-based anomaly methods~\cite{akoglu2015graph}.

\medskip\noindent\textbf{Deep Learning for Anomaly Detection.}\quad While shallow methods such as one-class classification models require explicit hand-crafted features, much of the recent work in anomaly detection leverages deep learning~\cite{Guansong:20}, which has shown remarkable capabilities in learning discriminative feature representations by extracting high-level features from data using multilayered neural networks. Ruff \textit{et al.}~\cite{ruff2018deep} develop a deep SVDD anomaly detection framework, which is basically an extension of the shallow one-class classification SVDD approach. The basic idea behind SVDD is to find the smallest hypersphere that contains all instances, except for some anomalies. Deep SVDD is an unsupervised learning model that learns to extract the common factors of variation of the data distribution by training a neural network while minimizing the volume of a hypersphere that encloses the network representations of the data. The centroid of the hypersphere is usually set to the mean of the feature representations learned by performing a single initial forward pass. In order to improve model performance, Ruff \textit{et al.}~\cite{ruff2019deep} propose Deep SAD, a generalization of the unsupervised Deep SVDD to the semi-supervised setting. The key difference between these two deep anomaly detection models is that the objective function of Deep SAD also includes a loss term for labeled normal and anomalous instances. The idea behind this loss term is to minimize (resp. maximize) the squared Euclidean distance between the labeled normal (resp. anomalous) instances and the hypersphere centroid. However, both Deep SVDD and Deep SAD suffer from the hypersphere collapse problem due to the learning of a trivial solution. In other words, the network's learned features tend to converge to the centroid of the hypersphere if no constraints are imposed on the architectures of the models. Cevikalp \textit{et al.}~\cite{Cevikalp2023Gap} considered the hypersphere centers as parameters that can be learned and updated according to the evolving deep feature representations. Another line of work uses deep generative models to address the anomaly detection problem~\cite{Donahue:17,Schlegl:19}. These generative networks are able to localize anomalies, particularly in images, by simultaneously training a generator and a discriminator, enabling the detection of anomalies on unseen data based on unsupervised training of the model on anomaly-free data~\cite{DiMattia:19}. However, the use of deep generative models in anomaly detection has been shown to be quite problematic and unintuitive, particularly on image data~\cite{Nalisnick:19}.

\medskip\noindent\textbf{Graph Convolutional Networks for Anomaly Detection.}\quad GCNs have recently become the de facto model for learning representations on graphs, achieving state-of-the-art performance in various application domains, including anomaly detection~\cite{ding2019deep,kumagai2020semi}. Ding~\textit{et al.}~\cite{ding2019deep} present an unsupervised graph anomaly detection framework using a GCN-based autoencoder. This approach leverages both the topological structure and nodal attributes, with an encoder that maps the attribute information into a low-dimensional feature space and a decoder that reconstructs the structure as well as the attribute information using the learned latent representations. The basic idea behind this GCN-based autoencoder is that the normal instances can be reconstructed with small errors, while anomalous instances are often reconstructed with large errors. However, methods based on reconstruction errors are prone to outliers and often require noise-free data for training. On the other hand, some of the main challenges associated with graph anomaly detection is the lack of labeled graph nodes (i.e. no information is available about which instances are actually anomalous and which ones are normal) and data imbalance, as abnormalities occur rarely and hence a tiny fraction of instances is expected to be anomalous. To circumvent these issues, Kumagai~\textit{et al.}~\cite{kumagai2020semi} propose two semi-supervised anomaly detection models using GCNs for learning latent representations. The first model uses only labeled normal instances, while the second one employs both labeled normal and anomalous instances. However, both models are trained to minimize the volume of a hypersphere that encloses the GCN-learned node embeddings of normal instances, and hence they also suffer from the hypersphere collapse problem. By contrast, our semi-supervised GFCN model does not suffer from the above mentioned issues. In addition to leveraging the graph structure and node attributes, GFCN learns from both labeled and unlabeled data in order to improve model performance.

\medskip\noindent\textbf{Graph Neural Networks with Skip Connections.}\quad Despite the success of GNN-based models in learning node representations, they are prone to over-smoothing, which can negatively impact their performance. Over-smoothing occurs when stacking multiple graph convolution layers causes node representations to become indistinguishable, leading to a loss of valuable information. To tackle the over-smoothing problem, several approaches that leverage skip connections have been proposed. Skip connections can be categorized into four main types: residual connections, initial connections, jumping connections, and dense connections~\cite{huang2022graph}. JK-Net~\cite{Keyulu:18} uses jumping knowledge network connections to connect each layer to the last one, maintaining the feature mappings in lower layers. APPNP~\cite{gasteiger2018combining}, which approximate PageRank with power iteration, uses initial connection by connecting each layer to the original feature matrix. By decoupling feature transformation and propagation, APPNP can aggregate information from multi-hop neighbors without increasing the number of layers in the network. GCNII~\cite{Chen2020GCNII} employs initial residual and identity mapping to mitigate the over-smoothing problem. At each layer, the initial residual constructs a skip connection from the input layer, while the identity mapping adds an identity matrix to the weight matrix. ResGCN~\cite{Li2019DeepGCN} is a residual graph convolutional network that extends the depth of GCNs by using residual/dense connections and dilated convolutions. In our proposed GFCN model, we apply a skip connection that reuses the initial node features at each layer with the goal of combining both the aggregated node neighborhood representation and the initial node representation. While most graph convolutions with skip connections are based on heuristics, our GFCN model is theoretically motivated by implicit fairing and its layerwise propagation rule is derived from the iterative solution of the implicit fairing equation via the Jacobi method.

\section{Preliminaries and Problem Statement}
We introduce our notation and present a brief background on spectral graph theory~\cite{Hamza2007IP,Emad2007GI,Emad2009ISIVP}, followed by our problem formulation of semi-supervised anomaly detection on graphs.

\medskip\noindent\textbf{Basic Notions.}\quad Consider a graph $\mathcal{G}=(\mathcal{V},\mathcal{E})$, where $\mathcal{V}=\{1,\ldots,N\}$ is the set of $N$ nodes and $\mathcal{E}\subseteq \mathcal{V}\times\mathcal{V}$ is the set of edges. The graph structure is encoded by an $N\times N$ adjacency matrix $\bm{A}=(\bm{A}_{ij})$ whose $(i,j)$-th entry is equal to the weight of the edge between neighboring nodes $i$ and $j$, and 0 otherwise. We also denote by $\bm{X}=(\bm{x}_{1},...,\bm{x}_{N})^{\T}$ an $N\times F$ feature matrix of node attributes, where $\bm{x}_{i}$ is an $F$-dimensional row vector for node $i$. This real-valued feature vector is often referred to as a graph signal, which assigns a value to each node in the graph.

\medskip\noindent\textbf{Spectral Graph Theory.}\quad The normalized Laplacian matrix is defined as
\begin{equation}
\bm{L}=\bm{I}-\bm{D}^{-\frac{1}{2}}\bm{A}\bm{D}^{-\frac{1}{2}},
\end{equation}
where $\bm{D}=\op{diag}(\bm{A}\bm{1})$ is the diagonal degree matrix, and $\bm{1}$ is an $N$-dimensional vector of all ones. Since the normalized Laplacian matrix is symmetric positive semi-definite, it admits an eigendecomposition given by $\bm{L}=\bm{U}\bm{\Lambda}\bm{U}^{\T}$, where $\bm{U}=(\bm{u}_1,\dots,\bm{u}_N)$ is an orthonormal matrix whose columns constitute an orthonormal basis of eigenvectors and $\bm{\Lambda}=\op{diag}(\lambda_1,\dots,\lambda_N)$ is a diagonal matrix comprised of the corresponding eigenvalues such that $0=\lambda_1\le\dots\le\lambda_N\le 2$. If $\mathcal{G}$ is a bipartite graph, then the spectral radius (i.e. largest absolute value of all eigenvalues) of the normalized Laplacian matrix is equal to 2. The normalized Laplacian matrix has eigenvalues in the range [0,2], which makes spectral graph analysis algorithms more stable and reliable compared to algorithms that use the unnormalized Laplacian matrix with eigenvalues that can be much larger. Moreover, scaling by the inverse square root of the degree matrix helps reduce the influence of highly connected nodes.

\medskip\noindent\textbf{Problem Statement.}\quad Anomaly detection aims at identifying anomalous instances, which do not conform to the expected pattern of other instances in a dataset. It differs from binary classification in that it distinguishes between normal and anomalous observations. Also, the distribution of anomalies is not usually known a priori.

Let $\mathcal{D}_{l}=\{(\bm{x}_i,y_i)\}_{i=1}^{N_l}$ be a set of labeled data points $\bm{x}_i\in\mathbb{R}^{F}$ and their associated known labels $y_i\in\{0,1\}$ with 0 and 1 representing ``normal'' and ``anomalous'' observations, respectively, and $\mathcal{D}_{u}=\{\bm{x}_i\}_{i=N_l+1}^{N_l+N_u}$ be a set of unlabeled data points, where $N_l+N_u=N$. Hence, each node $i$ can be labeled with a 2-dimensional one-hot encoding vector $\bm{y}_i=(y_{i},1-y_{i})$.

The goal of semi-supervised anomaly detection on graphs is to estimate the anomaly scores of the unlabeled graph nodes. Nodes with high anomaly scores are considered anomalous, while nodes with lower scores are deemed normal.

\section{Proposed Method}
In this section, we begin by succinctly summarizing some of the most common spectral filters on graphs. Then, we propose a graph convolutional network with skip connection using the concept of implicit fairing on graphs. In particular, we examine the main components of the proposed architecture and analyze the complexity of the model. In addition, we introduce an anomaly scoring function defined in terms of the weighted cross-entropy between the ground-truth labels of the graph test nodes and the model's predicted probabilities.

\subsection{Spectral Graph Filtering}
The idea of spectral filtering on graphs was first introduced in~\cite{Taubin:95} in the context of triangular mesh smoothing. The goal of spectral graph filtering is to use polynomial or rational polynomial filters defined as functions of the graph Laplacian (or equivalently its eigenvalues) in an effort to attenuate high-frequency noise corrupting the graph signal. These functions are usually referred to as frequency responses or transfer functions. While polynomial filters have finite impulse responses, their rational counterparts have infinite impulse responses. Despite the fact that the Laplacian matrix is commonly used in spectral graph theory, it does not, however, provide a natural way to normalize the frequency domain representation of a graph signal, which can lead to scaling and convergence issues in spectral graph filtering. In contrast, the normalized Laplacian matrix provides a way to normalize the frequency domain representation of a graph signal, which can improve the stability and convergence properties of spectral graph filtering. Specifically, the normalized Laplacian matrix is scaled by the inverse square root of the degree matrix, which helps normalize the contributions of each node's neighbors to the overall graph signal. Applying a spectral graph filter with transfer function $h$ on the graph signal $\bm{X}\in\mathbb{R}^{N\times F}$ yields
\begin{equation}
\bm{H}=h(\bm{L})\bm{X}=\bm{U}h(\bm{\Lambda})\bm{U}^{\T}\bm{X}=\bm{U}\op{diag}(h(\lambda_i))\bm{U}^{\T}\bm{X},
\end{equation}
where $\bm{H}$ is the filtered graph signal. However, this filtering process necessitates the computation of the eigenvalues and eigenvectors of the Laplacian matrix, which is prohibitively expensive for large graphs. To circumvent this issue, spectral graph filters are usually approximated using Chebyshev polynomials~\cite{Taubin:96,Hammond:11,Defferrard:16} or rational polynomials~\cite{Levie:18,Bianchi:19,Wijesinghe:19}.

\subsection{Implicit Fairing}
Graph fairing refers to the process of designing and computing smooth graph signals on a graph in order to filter out undesirable high-frequency noise while retaining the graph geometric features as much as possible. The implicit fairing method, which uses implicit integration of a diffusion process for graph filtering, has shown to allow for both efficiency and stability~\cite{Desbrun:99}. The implicit fairing filter is an infinite impulse response filter whose transfer function is given by $h_{s}(\lambda)=1/(1+s\lambda)$, where $s$ is a positive parameter. Hence, performing graph filtering with implicit fairing is equivalent to solving the following sparse linear system:
\begin{equation}
(\bm{I}+s\bm{L})\bm{H}=\bm{X},
\label{Eq:IF}
\end{equation}
which we refer to as implicit fairing equation. It is worth pointing out that this equation can also be obtained by minimizing the following objective function
\begin{equation}
\mathcal{J}(\bm{H}) =\frac{1}{2}\Vert\bm{H}-\bm{X}\Vert_{F}^{2}+\frac{s}{2}\op{tr}(\bm{H}^{\T}\bm{L}\bm{H}),
\end{equation}
where $\Vert\cdot\Vert_{F}$ and $\op{tr}(\cdot)$ denote the Frobenius norm and trace operator, respectively.

\begin{figure}[!htb]
\centering
\includegraphics[scale=0.62]{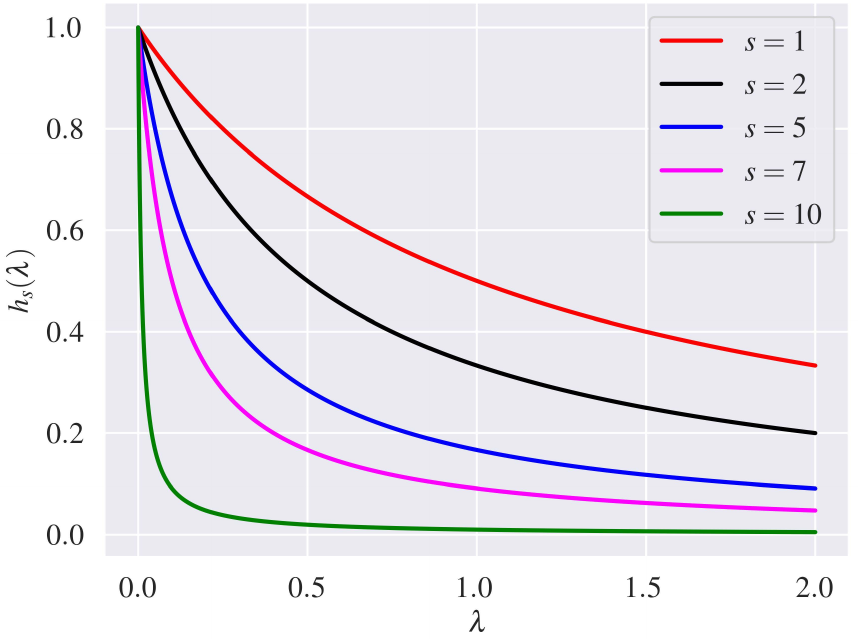}
\caption{Transfer function of the implicit fairing filter for various values of the scaling parameter.}
\label{Fig:TF}
\end{figure}

The implicit fairing filter enjoys several nice properties, including unconditional stability as $h_{s}(\lambda)$ is always in $[0,1]$, and also preservation of the average value (i.e. DC value or centroid) of the graph signal as $h_{s}(0)=1$ for all $s$. As shown in Figure~\ref{Fig:TF}, the higher the value of the scaling parameter, the closer the implicit fairing filter becomes to the ideal low-pass filter.

\subsection{Spectral Analysis}
The matrix $\bm{I}+s\bm{L}$ is symmetric positive definite with minimal eigenvalue equal to 1 and maximal eigenvalue bounded from above by $1+2s$. Hence, the condition number $\kappa$ of $\bm{I}+s\bm{L}$ satisfies
\begin{equation}
\kappa\le 1+2s,
\end{equation}
where $\kappa$, which is defined as the ratio of the maximum to minimum stretching, is also equal to the maximal eigenvalue of $\bm{I}+s\bm{L}$. Intuitively, the condition number measures how much can a change (i.e. small perturbation) in the right-hand side of the implicit fairing equation affects the solution. In fact, it can be readily shown that the resulting relative change in the solution of the implicit fairing equation is bounded from above by the condition number multiplied by the relative change in the right-hand side.

\subsection{Iterative Solution}
One of the simplest iterative techniques for solving a matrix equation is Jacobi's method, which uses matrix splitting. Since the matrix $\bm{I}+s\bm{L}$ can be split into the sum of a diagonal matrix and an off-diagonal matrix
\begin{equation}
\bm{I}+s\bm{L}=(1+s)\bm{I}-s\bm{D}^{-1/2}\bm{A}\bm{D}^{-1/2},
\end{equation}
the implicit fairing equation can then be solved iteratively using the Jacobi method as follows:
\begin{equation}
\begin{split}
\bm{H}^{(t+1)}
&=\bm{D}^{-1/2}\bm{A}\bm{D}^{-1/2}\bm{H}^{(t)}\bm{\Theta}_{s}+\bm{X}\widetilde{\bm{\Theta}}_{s},
\end{split}
\label{Eq:ISJ}
\end{equation}
where $\bm{\Theta}_{s}=\op{diag}(s/(1+s))$ and $\widetilde{\bm{\Theta}}_{s}=\op{diag}(1/(1+s))$ are $F\times F$ diagonal matrices, each of which has equal diagonal entries, and $\bm{H}^{(t)}$ is the $t$-th iteration of $\bm{H}$. Since the spectral radius of the normalized adjacency matrix is equal to 1, it follows that the spectral radius of the Jacobi's iteration matrix
\begin{equation}
\bm{C}=\frac{s}{1+s}\bm{D}^{-1/2}\bm{A}\bm{D}^{-1/2},
\end{equation}
is equal to $s/(1+s)$, which is always smaller than 1. Hence, the convergence of the iterative method given by Eq. \eqref{Eq:ISJ} holds.

\subsection{Graph Fairing Convolutional Network}
At the core of graph representation learning is the concept of propagation rule, which determines how information is passed between nodes in a graph. It involves updating the current node features by aggregating information from their neighboring nodes, followed by a non-linear activation function to produce an updated representation for the node. Inspired by the Jacobi iterative solution of the implicit fairing equation, we propose a multi-layer graph fairing convolutional network (GFCN) with the following layer-wise propagation rule:
\begin{equation}
\bm{H}^{(\ell+1)}=\sigma(\bm{D}^{-1/2}\bm{A}\bm{D}^{-1/2}\bm{H}^{(\ell)}\bm{\Theta}^{(\ell)}
+\bm{X}\widetilde{\bm{\Theta}}^{(\ell)}),
\label{Eq:IS}
\end{equation}
where $\bm{\Theta}^{(\ell)}$ and $\widetilde{\bm{\Theta}}^{(\ell)}$ are learnable weight matrices, $\sigma(\cdot)$ is an element-wise activation function, $\bm{H}^{(\ell)}\in\mathbb{R}^{N\times F_{\ell}}$ is the input feature matrix of the $\ell$-th layer with $F_{\ell}$ feature maps for $\ell=0,\dots,L-1$. The input of the first layer is the initial feature matrix $\bm{H}^{(0)}=\bm{X}$.

\begin{figure}[!t]
\centering
\includegraphics[scale=.5]{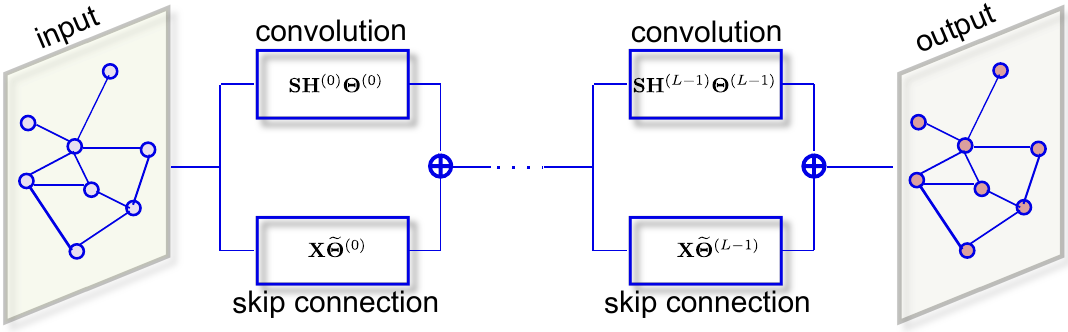}
\caption{Schematic layout of the proposed GFCN architecture. Each block comprises a graph convolution and a skip connection, followed by an activation function, where $\bm{S}$ denotes the normalized adjacency matrix. The GFCN model takes as input the adjacency matrix $\bm{A}$ and initial feature matrix $\bm{X}=\bm{H}^{(0)}$. At each layer, a node aggregates information from its neighboring nodes and the initial feature matrix through skip connection. The aggregated information is then transformed using learnable weight matrices. The resulting representation is then passed to the next layer for further propagation. Finally, the output is the latent graph representation $\bm{H}^{(L)}$ from the last network layer.}
\label{Fig:GFCN}
\end{figure}

Note that in addition to performing graph convolution, which essentially averages the features of the immediate (i.e. first-order or 1-hop) neighbors of nodes, the layer-wise propagation rule of GFCN also applies a skip connection that reuses the initial node features, as illustrated in Figure~\ref{Fig:GFCN}. In other words, GFCN combines both the aggregated node neighborhood representation and the initial node representation, hence memorizing information across distant nodes. While most of the existing graph convolutions with skip connections~\cite{Keyulu:18,Bianchi:19} are based on heuristics, our graph fair convolution is theoretically motivated by implicit fairing and derived directly from the Jacobi iterative method. It is important to mention that the convolution operation in the proposed propagation rule involves aggregating information from a node's neighbors and combining it with the node's own features to produce a new set of features. As a result, the normalized adjacency matrix $\bm{D}^{-1/2}\bm{A}\bm{D}^{-1/2}$ is particularly useful in this context because it provides a way to normalize the aggregation of neighbor features, which helps avoid the problem of over-reliance on highly connected nodes. This normalization also helps make the convolution operation more stable and better conditioned, which can improve the convergence and generalization performance of our mode. The normalized adjacency matrix considers both the number of neighbors connected to a node and the number of neighbors connected to each of those neighboring nodes.

In contrast to GCN-based methods which use a first-order approximation of spectral graph convolutions and a renormalization trick in their layer-wise propagation rules to avoid numerical instability, our GFCN model does not require any renormalization as the spectral radius of the normalized adjacency matrix is equal to 1, while still maintaining the key property of convolution as a neighborhood aggregation operator. Hence, repeated application of the GFCN's layer-wise propagation rule provides a computationally efficient convolutional process, leading to numerical stability and avoidance of exploding/vanishing gradients. It should also be pointed out that similar to GCN, the proposed GFCN model also does not require explicit computation of the Laplacian eigenbasis. In addition, the aggregation of GFCN with skip connections between the initial feature matrix and each hidden layer does not require filtered learning of the hidden layers. Skip connections are a mechanism that allows our model to directly pass on the information from the initial feature matrix (input) to the deeper hidden layers, thereby helping to retain the original node features to prevent loss of important information during the aggregation process.

\subsection{Model Prediction}
\begin{figure}[!h]
\centering
\includegraphics[scale=0.55]{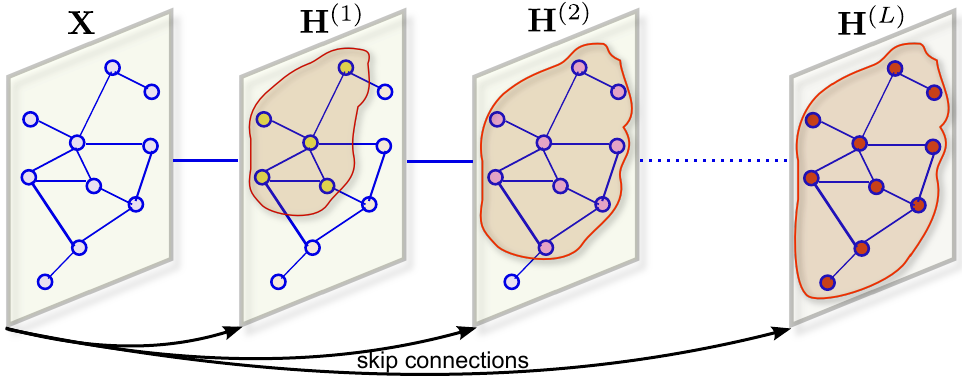}
\caption{Illustration of the GFCN aggregation scheme with skip connections.}
\label{Fig:GFCNskip}
\end{figure}

The layer-wise propagation rule of GFCN is basically a node embedding transformation that projects both the input $\bm{H}^{(\ell)}\in\mathbb{R}^{N\times F_{\ell}}$ into a trainable weight matrix $\bm{\Theta}^{(\ell)}\in\mathbb{R}^{F_{\ell}\times F_{\ell+1}}$ and the initial feature matrix $\bm{X}$ into the skip-connection weight matrix $\widetilde{\bm{\Theta}}^{(\ell)}\in\mathbb{R}^{F\times F_{\ell+1}}$, with $F_{\ell +1}$ feature maps. Then, a point-wise activation function $\sigma(\cdot)$ such as $\text{ReLU}(\cdot)=\max(0,\cdot)$ is applied to obtain the output node embedding matrix. The aggregation scheme of GFCN is depicted in Figure~\ref{Fig:GFCNskip}. Note that GFCN uses skip connections between the initial feature matrix and each hidden layer. Skip connections not only allow the model to carry over information from the initial node attributes, but also help facilitate training of multi-layer networks.

The embedding $\bm{H}^{(L)}$ of the last layer of GFCN contains the final output node embeddings, which can be used as input for downstream tasks such as graph classification, clustering, visualization, link prediction, recommendation, node classification, and anomaly detection. Since anomaly detection can be posed as a binary classification problem, we apply the point-wise softmax function (i.e. sigmoid in the binary case) to obtain an $N\times 2$ matrix of predicted labels for graph nodes
\begin{equation}
\hat{\bm{Y}}=(\hat{\bm{y}}_{1},\dots,\hat{\bm{y}}_{N})^{\T}=\text{softmax}(\bm{H}^{(L)}),
\end{equation}
where $\hat{\bm{y}}_{i}=(\hat{y}_{i},1-\hat{y}_{i})$ is a two-dimensional raw vector of predicted probabilities, with $\hat{y}_{i}$ the probability that the network associates the $i$-th node with one of the two classes (i.e. 1 for anomalous and 0 for normal).

\subsection{Model Complexity}
For simplicity, we assume the feature dimensions are the same for all layers, i.e. $F_{\ell}=F$ for all $\ell$, with $F \ll N$. The time complexity of an $L$-layer GFCN is $\mathcal{O}(L\Vert\bm{A}\Vert_{0}F+LNF^2)$, where $\Vert\bm{A}\Vert_{0}$ denotes the number of non-zero entries of the sparse adjacency matrix. Note that multiplying the normalized adjacency matrix with an embedding costs $\mathcal{O}(\Vert\bm{A}\Vert_{0}F)$ in time, while multiplying an embedding with a weight matrix costs $\mathcal{O}(NF^2)$. Also, multiplying the initial feature matrix by the skip-connection weight matrix costs $\mathcal{O}(NF^2)$.

For memory complexity, an $L$-layer GFCN requires $\mathcal{O}(LNF+LF^2)$ in memory, where $\mathcal{O}(LNF)$ is for storing all embeddings and $\mathcal{O}(LF^2)$ is for storing all layer-wise weight matrices.

Therefore, our proposed GFCN model has the same time and memory complexity as GCN, albeit GFCN takes into account distant graph nodes for improved learned node representations.

\subsection{Model Training}
The parameters (i.e. weight matrices for different layers) of the proposed GFCN model for semi-supervised anomaly detection are learned by minimizing the following regularized loss function
\begin{equation}
\mathcal{L} =\frac{1}{N_l}\sum_{i=1}^{N_l}\mathcal{C}_{\alpha}(\bm{y}_i,\hat{\bm{y}}_i)
+\frac{\beta}{2}\sum_{\ell=0}^{L-1}\left(\Vert\bm{\Theta}^{(\ell)}\Vert_{F}^2
+\Vert\widetilde{\bm{\Theta}}^{(\ell)}\Vert_{F}^2\right),
\end{equation}
where $\mathcal{C}_{\alpha}(\bm{y}_i,\hat{\bm{y}}_i)$ is the weighted cross-entropy given by
\begin{equation}
\mathcal{C}_{\alpha}(\bm{y}_i,\hat{\bm{y}}_i)=-\alpha\,y_{i}\log\hat{y}_i - (1-y_{i})\log(1-\hat{y}_i),
\end{equation}
which measures the dissimilarity between the one-hot encoding vector $\bm{y}_i$ of the $i$th node and the corresponding vector $\hat{\bm{y}}_i$ of predicted probabilities. This dissimilarity decreases as the value of the predicted probability approaches the ground-truth label. The weight parameter $\alpha$ adjusts the importance of the positive labels by assigning more weight to the anomalous class, while the parameter $\beta$ controls the importance of the regularization term, which is added to prevent overfitting. The regularization term is the sum of the squared elements of the learnable weight matrices for each layer. It is important to mention that we only use the normal class labeled instances to train our model.

We optimize our model using the Adam optimizer~\cite{Kingma2015Adam}, which is a modified version of Stochastic Gradient Descent (SGD) that uses adaptive moment estimation. The intuition behind the use of the weighted cross-entropy loss function is to assign a higher weight to the anomalous nodes than to the normal nodes, so that the model is encouraged to correctly identify the anomalous nodes even if they are rare and overshadowed by the large number of normal nodes. The regularization term, on the other hand, penalizes large weight values in the learnable weight matrices, which helps to reduce the complexity of the model and improve its generalization performance. The strength of the regularization is controlled by the value of the hyperparameter $\beta$, which is tuned using grid search.

The weight matrices of our GFCN model are initialized randomly with small values using a normal distribution to ensure that the variance of the activations and gradients is roughly the same across all layers of the network. During training, the optimizer adjusts the weight matrices to minimize the regularized loss function. The training process involves choosing the hyperparameters, computing the regularized weighted cross-entropy loss, feeding forward and backpropagating the inputs, and updating the weight matrices using the Adam optimizer. This process is repeated for multiple epochs until the model converges or the validation loss does not decrease after a specified number of consecutive epochs.

\subsection{Model Inference}
Once the model is trained, we can use the weighted cross-entropy errors to assess the abnormality of nodes. To this end, we define the anomaly score of the $i$th test node as
\begin{equation}
s_{i} = \mathcal{C}_{\alpha}(\bm{y}_i,\hat{\bm{y}}_i).
\end{equation}
Since the range of the weighted cross-entropy is $[0,\infty]$ (e.g. infinite value when $y_{i}=1$ and $\hat{y}_i=0$), we apply min-max normalization to bring all anomaly scores into the range [0,1] as follows:
\begin{equation}
\tilde{s}_{i} = \frac{s_{i}-s_{\min}}{s_{\max}-s_{\min}},
\end{equation}
where $s_{\min}$ and $s_{\max}$ are the minimum and maximum, respectively, of the anomaly scores in the test set. Nodes with scores larger than a certain threshold are considered anomalies. Hence, we can compute a ranked list of anomalies according to their normalized anomaly scores. In other words, we compute the anomaly scores of each node in the test set, and then the top-$r$ nodes with higher scores are identified as anomalies for a user-specified value of $r$.

\section{Experiments}
In this section, we conduct extensive experiments to assess the performance of the proposed anomaly detection framework in comparison with state-of-the-art methods on several benchmark datasets. The source code to reproduce the experimental results is made publicly available on GitHub\footnote{https://github.com/MahsaMesgaran/GFCN}.

\subsection{Datasets}
We demonstrate and analyze the performance of the proposed model on three citation networks: Cora, Citeseer, and Pubmed~\cite{Sen:08}, and two co-purchase graphs: Amazon Photo and Amazon Computers~\cite{shchur2018pitfalls}. The summary descriptions of these benchmark datasets are as follows:
\begin{itemize}
\item Cora is a citation network dataset consisting of 2708 nodes representing scientific publications and 5429 edges representing citation links between publications. All publications are classified into 7 classes (research topics). Each node is described by a binary feature vector indicating the absence/presence of the corresponding word from the dictionary, which consists of 1433 unique words.

\item Citeseer is a citation network dataset composed of 3312 nodes representing scientific publications and 4723 edges representing citation links between publications. All publications are classified into 6 classes (research topics). Each node is described by a binary feature vector indicating the
absence/presence of the corresponding word from the dictionary, which consists of 3703 unique words.

\item Pubmed is a citation network dataset containing 19717 scientific publications pertaining to diabetes and 44338 edges representing citation links between publications. All publications are classified into 3 classes. Each node is described by a TF/IDF weighted word vector from the dictionary, which consists of 500 unique words.

\item Amazon Computers and Amazon Photo datasets are co-purchase graphs~\cite{shchur2018pitfalls}, where nodes represent goods and edges indicate that two goods are frequently bought together. The node features are bag-of-words encoded product reviews, while the class labels are given by the product category.

\item ogbn-arxiv dataset is a large-scale graph dataset from open graph benchmark (OGB) representing the citation network between all computer science (CS) arXiv papers. In this dataset over 169k nodes and 1.1m edges are contained. Each node has a 128-dimensional feature vector obtained by averaging the embeddings of words in the article's title and abstract.
\end{itemize}

Since there is no ground truth of anomalies in these datasets, we employ the commonly-used protocol~\cite{kumagai2020semi} in anomaly detection by treating the smallest class for each dataset as the anomaly class and the remaining classes as the normal class. Dataset statistics are summarized in Table~\ref{Tab:Dataset statistics}, where anomaly rate refers to the percentage of abnormalities in each dataset. For all datasets, we only use the normal class labeled instances to train the model.

\begin{table}[!htb]
\setlength{\tabcolsep}{1.8pt}
\caption{Summary statistics of datasets.}
\small
\medskip
\centering
\begin{tabular}{lrrrrc}
\toprule[1pt]
\textbf{Dataset}& \textbf{Nodes}& \textbf{Edges}&\textbf{Features}& \textbf{Classes}&\textbf{Anomaly Rate (\%)}\\
\midrule[.8pt]
\textbf{Cora}& 2708& 5278& 1433& 7& 0.06\\
\textbf{Citeseer}& 3327& 4732& 3703 &6& 0.07\\
\textbf{Pubmed}& 19717& 44338& 500 &3& 0.21\\
\textbf{Photo}& 7487& 119043& 745& 8& 0.04\\
\textbf{Computers}& 13381& 245778& 767& 10& 0.02\\
\textbf{ogbn-arxiv}& 169343&1166243&128&40&0.02\\
\bottomrule[1pt]
\end{tabular}
\label{Tab:Dataset statistics}
\end{table}

\subsection{Baseline Methods}
We evaluate the performance of the proposed method against various baselines, including one-class support vector machines (OC-SVMs)~\cite{Scholkopf:01}, imbalanced vertex diminished (ImVerde)~\cite{wu2018imverde}, one-class deep support vector data description (Deep SVDD)~\cite{ruff2018deep}, deep anomaly detection on attributed networks (Dominant)~\cite{ding2019deep}, one-class deep semi-supervised anomaly detection (Deep SAD)~\cite{ruff2019deep}, graph convolutional networks (GCNs)~\cite{Kipf:17}, GCN-based anomaly detection (GCN-N and GCN-AN)~\cite{kumagai2020semi}, graph random neural networks (GRAND)~\cite{feng2020graph}, semi-Supervised node classification on graph with few labels via non-parametric distribution assignment (GraFN)~\cite{lee2022grafn}, and re-weighting the influence of labeled nodes (ReNode)~\cite{chen2021topology}. For baselines, we mainly consider methods that are closely related to GFCN and/or the ones that are state-of-the-art anomaly detection frameworks. A brief description of these strong baselines can be summarized as follows:
\begin{itemize}
\item \textbf{OC-SVM}~\cite{Scholkopf:01} is an unsupervised one-class anomaly detection technique, which learns a discriminative hyperplane boundary around the normal instances using support vector machines by maximizing the distance from this hyperplane to the origin of the high-dimensional feature space.
\item \textbf{ImVerde}~\cite{wu2018imverde} is a semi-supervised graph representation learning technique for imbalanced graph data based on a variant of random walks by adjusting the transition probability each time a graph node is visited by the random particle.
\item \textbf{Deep SVDD}~\cite{ruff2018deep} is an unsupervised anomaly detection method, inspired by kernel-based one-class classification and minimum volume estimation, which learns a spherical, instead of a hyperplane, boundary in the feature space around the data using support vector data description. It trains a deep neural network while minimizing the volume of a hypersphere that encloses the network embeddings of the data. Normal instances fall inside the hypersphere, while anomalies fall outside.
\item \textbf{Dominant}~\cite{ding2019deep} is a deep autoencoder based on GCNs for unsupervised anomaly detection on attributed graphs. It employs an objective function defined as a convex combination of the reconstruction errors of both graph structure and node attributes. These learned reconstruction errors are then used to assess the abnormality of graph nodes.
\item \textbf{Deep SAD}~\cite{ruff2019deep} is a semi-supervised anomaly detection technique, which generalizes the unsupervised Deep SVDD approach to the semi-supervised setting by incorporating a new term for labeled training data into the objective function. The weights of the Deep SAD network are initialized using an autoencoder pre-training mechanism.
\item \textbf{GCN}~\cite{Kipf:17} is a deep graph neural network for semi-supervised learning of graph representations, encoding both local graph structure and attributes of nodes. It is an efficient extension of convolutional neural networks to graph-structured data, and uses a graph convolution that aggregates and transforms the feature vectors from the local neighborhood of a graph node.
\item \textbf{GCN-N} and \textbf{GCN-AN}~\cite{kumagai2020semi} are GCN-based, semi-supervised anomaly detection frameworks, which rely on minimizing the volume of a hypersphere that encloses the node embeddings to detect anomalies. Node embeddings placed inside and outside this hypersphere are deemed normal and anomalous, respectively. GCN-N uses only normal label information, while GCN-AN uses both anomalous and normal label information.
\item \textbf{GRAND}~\cite{feng2020graph} is a semi-supervised learning on graphs when labeled nodes are scarce. This technique relies on a random propagation strategy to perform graph data augmentation and employs consistency regularization to optimize prediction consistency of unlabeled nodes across different data augmentations.
\item \textbf{GraFN}~\cite{lee2022grafn} is a semi-supervised node representation learning for graphs with few labeled nodes. This technique exploits the self-supervised loss to ensure nodes that belong to the same class to be grouped together on differently augmented graphs. GraFN randomly samples support nodes from the labeled nodes and anchor nodes from the entire graph, and non-parametrically compute two predicted class distributions from two augmented graphs based on the anchor supports similarity.
\item \textbf{ReNode}~\cite{chen2021topology} is a semi-supervised node classification technique addressing the the topology-imbalance node representation learning as a graph specific imbalance learning problem. To measure the degree of topology imbalance, a conflict detection-based metric, Totoro, is used to locate node position. ReNode adjusts the training weights of labeled nodes based on their topological positions.
\end{itemize}

\subsection{Evaluation Metric}
In order to evaluate the performance of our proposed framework against the baseline methods, we use AUC, the area under the receiving operating characteristic (ROC) curve, as a metric. AUC summarizes the information contained in the ROC curve, which plots the true positive rate versus the false positive rate, at various thresholds~\cite{Pickup16IJCV,Biasotti16VC}. Larger AUC values indicate better performance at distinguishing between anomalous and normal instances. An uninformative anomaly detector has an AUC equal to 50\% or less. An AUC of 50\% corresponds to a random detector (i.e. for every correct prediction, the next prediction will be incorrect), whereas an AUC score smaller than 50\% indicates that a detector performs worse than a random detector.

\subsection{Implementation Details}
For fair comparison, we implement the proposed method and baselines in PyTorch using the PyTorch Geometric library. Following common practices for evaluating performance of GCN-based models~\cite{Kipf:17,kumagai2020semi}, we train our 2-layer GFCN model for 100 epochs using the Adam optimizer~\cite{Kingma2015Adam} with a learning rate of 0.1. We tune the latent representation dimension by hand, and set it to 128. The hyperparameters $\alpha$ and $\beta$ are chosen via grid search with cross-validation over the sets $\{10^{-4}, 10^{-3},\dots,1\}$ and $\{2,3,\dots,10\}$, respectively. We tune hyperparameters using the validation set, and terminate training if validation loss does not decrease after 10 consecutive epochs. For each dataset, we consider the settings where 2.5\%, 5\% and 10\% of instances are labeled, and we compute the average and standard deviation of test AUCs over ten runs.

\subsection{Anomaly Detection Performance}
Tables~\ref{Tab:AUC2.5}-\ref{Tab:AUC10} present the anomaly detection results on the five datasets. The best results are highlighted as bold. For each dataset, we report the AUC averaged over 10 runs as well as the standard deviation, at various ratios of labeled instances. As can be seen, our GFCN model consistently achieves the best performance on all datasets, except in the case of the Cora dataset when 10\% of instances are labeled. In that case, GCN-AN yields a marginal improvement of 0.7\% over GFCN, despite the fact that GCN-AN is trained on both normal and anomalous instances, whereas our model is trained only on normal instances. In addition, ImVerde, Deep SAD, GCN and GCN-AN all perform reasonably well on all datasets at various levels of label rates, but we find that GFCN outperforms these baselines on almost all datasets, while being considerably simpler. An AUC score of 50\% or less indicates that the baseline is an uninformative anomaly detector.

On the Amazon Computers dataset, Table~\ref{Tab:AUC2.5} shows that the proposed GFCN approach performs on par with GCN-AN, but outperforms all baselines on the other four datasets. In particular, GFCN yields 16.2\% and 15.7\% performance gains over Deep SAD on the Cora and Photo datasets, respectively. These gains are consistent with the results shown in Tables~\ref{Tab:AUC5}-\ref{Tab:AUC10}. We argue that the better performance of GFCN over GCN-N and Deep SAD is largely attributed to the fact that our model does not suffer from the hypersphere collapse problem. Interestingly, the performance gains are particularly higher at the lower label rate 2.5\%, confirming the usefulness of semi-supervised learning in that it improves model performance by leveraging unlabeled data. Another interesting observation is that in general both GCN-AN and GCN-N yield relatively high AUC standard deviations compared to our GFCN model, indicating that our model has less variability than these two strong baselines.

Lastly, we examined the training times (in seconds) for GFCN on Cora when $10\%$ of all instances were labeled. We also recorded the training times of GFCN, GCN-AN, GCN-N, and GCN on Cora for which we obtained 3.19, 4.12, 2.31, and 2.06 seconds, respectively. Since GFCN uses the skip connection, it took more training time than GCNs. However, the experiment shows the proposed method could learn the abnormalities fast enough.

\begin{table*}[!htb]
\caption{Test AUC (\%) averaged over 10 runs when 2.5\% of instances are labeled. We also report the standard deviation. Boldface numbers indicate the best anomaly detection performance.}
\small
\medskip
\centering
\begin{tabular}{l*{6}{c}}
\toprule[1pt]
& \multicolumn{6}{c}{\textbf{Dataset}}\\
\cmidrule(lr){2-7}
\textbf{Method} & Cora &Citeseer &	Pubmed & Photo & Computers&ogbn-arxiv\\
\midrule[.8pt]
OC-SVM~\cite{Scholkopf:01} & 50.0$\pm$0.1&	 50.6$\pm$0.4&	  68.9$\pm$0.9&	 51.9$\pm$0.6&	47.3$\pm$0.7&-\\
ImVerde~\cite{wu2018imverde} & 85.9$\pm$6.1&	 60.3$\pm$6.5&	 94.3$\pm$0.5&	 89.1$\pm$1.4&	98.5$\pm$0.7&-\\
Deep SVDD~\cite{ruff2018deep} &  69.6$\pm$6.5&	 55.3$\pm$1.6&	 73.7$\pm$6.3&	 52.3$\pm$1.4&	 46.6$\pm$1.5&- \\
Dominant~\cite{ding2019deep} &  52.3$\pm$0.9&	 53.9$\pm$0.6&	 50.8$\pm$0.4&	 38.1$\pm$0.4&	46.8$\pm$1.2&- \\
Deep SAD~\cite{ruff2019deep} & 72.7$\pm$6.0&	 53.8$\pm$2.9&	 91.3$\pm$2.4&	 81.9$\pm$5.7&	92.2$\pm$2.5&-\\
GCN~\cite{Kipf:17} &  84.9$\pm$6.9&	 60.9$\pm$6.0&	 96.2$\pm$0.1&	90.1$\pm$2.5&	 98.1$\pm$0.3& 51.2$\pm$0.1\\
GCN-AN~\cite{kumagai2020semi} &  88.8$\pm$5.4&	 65.6$\pm$4.7&	 95.6$\pm$0.3&	95.4$\pm$1.8&	\textbf{98.8}$\pm$0.3&-\\
GCN-N~\cite{kumagai2020semi} &  62.6$\pm$9.9&	56.0$\pm$4.4&	76.5$\pm$4.2&	55.1$\pm$11&	56.9$\pm$5.1&-\\
GRAND~\cite{feng2020graph} & 81.3$\pm$8.1&56.4$\pm$4.5&89.6$\pm$1.2&88.7$\pm$2.1&91.7$\pm$5.5&50.7$\pm$1.3\\
GraFN~\cite{lee2022grafn} & 58.2$\pm$5.8&56.3$\pm$3.2&81.4$\pm$0.9&97.4$\pm$1.5&96.1$\pm$3.9&57.3$\pm$1.4\\
ReNode~\cite{chen2021topology} & 68.4$\pm$6.8&55.1$\pm$2.0&82.3$\pm$1.5&84.9$\pm$3.4&96.3$\pm$9.2&50.2$\pm$3.1\\
\midrule[.8pt]
\textbf{GFCN}& \textbf{93.9}$\pm$2.3&	\textbf{68.3}$\pm$1.1 &	\textbf{96.3}$\pm$0.1&	\textbf{97.6}$\pm$0.5&	 \textbf{98.8}$\pm$0.4&\textbf{60.3}$\pm$0.1\\
\bottomrule[1pt]
\end{tabular}
\label{Tab:AUC2.5}
\end{table*}

\begin{table*}[!htb]
\caption{Test AUC (\%) averaged over 10 runs when 5\% of instances are labeled. We also report the standard deviation. Boldface numbers indicate the best anomaly detection performance.}
\small
\medskip
\centering
\begin{tabular}{l*{6}{c}}
\toprule[1pt]
& \multicolumn{6}{c}{\textbf{Dataset}}\\
\cmidrule(lr){2-7}
\textbf{Method}& Cora&	Citeseer&	Pubmed&	Photo&	Computers&ogbn-arxiv\\
\midrule[.8pt]
OC-SVM~\cite{Scholkopf:01} &  50.2$\pm$0.1&	 50.7$\pm$0.5&	  71.0$\pm$1.1&	 51.9$\pm$0.6&	47.2$\pm$0.8&-\\
ImVerde~\cite{wu2018imverde}  & 91.1$\pm$3.2 & 64.5$\pm$4.5 & 94.9$\pm$0.5 & 92.2$\pm$1.2 & 98.6$\pm$0.6&-  \\
Deep SVDD~\cite{ruff2018deep} &	58.3$\pm$2.8 &	 56.0$\pm$0.8 &	 79.9$\pm$3.4 &	 52.3$\pm$0.8 &	47.0$\pm$1.4&- \\
Dominant~\cite{ding2019deep} &	 52.5$\pm$0.9 &	53.9$\pm$0.7 &	 53.0$\pm$0.5 &	38.1$\pm$0.5 &	 47.2$\pm$1.2&- \\
Deep SAD~\cite{ruff2019deep} &	 72.4$\pm$5.7 &	 61.1$\pm$4.2 &	 91.7$\pm$1.7 &	89.8$\pm$3.1 &	92.8$\pm$2.6&- \\
GCN~\cite{Kipf:17} &	 89.2$\pm$7.6 &	 63.9$\pm$4.7 &	 96.6$\pm$0.1 &	92.3$\pm$1.2 &	98.3$\pm$0.3& 53.0$\pm$0.1\\
GCN-AN~\cite{kumagai2020semi} &	 91.8$\pm$5.4 &	 68.3$\pm$3.8 &	 96.2$\pm$0.2 &	97.0$\pm$0.7 &	99.1$\pm$0.3&- \\
GCN-N~\cite{kumagai2020semi} &	 67.1$\pm$5.8 &	57.4$\pm$3.1 &	76.1$\pm$4.8 &	56.2$\pm$9.6 &	58.2$\pm$5.8&- \\
GRAND~\cite{feng2020graph} & 84.7$\pm$9.1&56.8$\pm$3.1&87.8$\pm$1.4&98.4$\pm$4.1&97.8$\pm$5.8&60.2$\pm$5.3\\
GraFN~\cite{lee2022grafn} & 61.3$\pm$12.6&56.1$\pm$4.3&83.0$\pm$6.1&98.2$\pm$5.6&96.6$\pm$7.9&59.8$\pm$6.1\\
ReNode~\cite{chen2021topology} & 69.9$\pm$3.3&56.1$\pm$1.6&82.7$\pm$1.1&85.7$\pm$2.1&97.5$\pm$2.1&60.1$\pm$1.9\\
\midrule[.8pt]
\textbf{GFCN} &	 \textbf{92.2}$\pm$2.3 &	\textbf{71.3}$\pm$1.3 &	\textbf{96.7}$\pm$0.1&	\textbf{98.8}$\pm$0.4&	 \textbf{99.3}$\pm$0.4&\textbf{61.1}$\pm$0.4\\
\bottomrule[1pt]
\end{tabular}
\label{Tab:AUC5}
\end{table*}

\begin{table*}[!htb]
\caption{Test AUC (\%) averaged over 10 runs when 10\% of instances are labeled. We also report the standard deviation. Boldface numbers indicate the best anomaly detection performance.}
\small
\medskip
\centering
\begin{tabular}{l*{6}{c}}
\toprule[1pt]
& \multicolumn{6}{c}{\textbf{Dataset}}\\
\cmidrule(lr){2-7}
\textbf{Method}& Cora &	Citeseer & Pubmed & Photo & Computers & ogbn-arxiv\\
\midrule[.8pt]
OC-SVM~\cite{Scholkopf:01} &  51.8$\pm$1.7&	 51.1$\pm$0.9&	  73.1$\pm$0.8&	 51.7$\pm$0.9&	47.4$\pm$0.8&-\\
ImVerde~\cite{wu2018imverde}  &	 94.5 $\pm$2.1   &	 68.5 $\pm$6.9   &	 95.5 $\pm$0.4   &	 92.8 $\pm$1.0   &	99.1 $\pm$0.4 &-  \\
Deep SVDD~\cite{ruff2018deep} &	59.8 $\pm$4.8 &	56.7 $\pm$1.7 &	 93.2 $\pm$1.0 &	 53.1 $\pm$0.6 &	47.5 $\pm$1.0&- \\
Dominant~\cite{ding2019deep} &	 52.6 $\pm$0.9 &	53.9 $\pm$0.8 &	 53.1 $\pm$0.4 &	38.1 $\pm$0.7 &	 46.6 $\pm$1.9 &-\\
Deep SAD~\cite{ruff2019deep} &	 72.9 $\pm$3.3 &	 62.4 $\pm$4.4 &	96.7 $\pm$0.1 &	89.5 $\pm$1.9 &	93.5 $\pm$1.9 &-\\
GCN~\cite{Kipf:17} &	 94.5 $\pm$4.3 &	 68.6 $\pm$3.3 &	 96.7 $\pm$0.1 &	93.3 $\pm$0.8 &	98.3 $\pm$0.3& 53.9$\pm$0.3\\
GCN-AN~\cite{kumagai2020semi} &	 \textbf{95.4} $\pm$2.7 &	 72.9 $\pm$4.3 &	 96.6 $\pm$0.1 &	97.9 $\pm$0.3 &	99.1 $\pm$0.3&- \\
GCN-N~\cite{kumagai2020semi} &	72.3 $\pm$7.0 &	60.1 $\pm$2.1 &	73.4 $\pm$5.7 &	53.6 $\pm$3.9 &	58.5 $\pm$4.6&- \\
GRAND~\cite{feng2020graph} & 86.5 $\pm$1.3&57.8$\pm$6.0&88.8$\pm$1.1&94.6$\pm$0.7&93.1$\pm$0.2&59.9$\pm$0.5\\
GraFN~\cite{lee2022grafn} & 78.7$\pm$7.4&60.3$\pm$8.1&83.1$\pm$6.3&98.7$\pm$4.0&97.2$\pm$1.7 & 62.6$\pm$3.3\\
ReNode~\cite{chen2021topology} & 68.3$\pm$4.2&54.6$\pm$0.8&83.4$\pm$1.4 & 90.4$\pm$2.3&99.0$\pm$0.3 & 61.2$\pm$1.8\\
\midrule[.8pt]
\textbf{GFCN} &	94.7 $\pm$1.0 &	 \textbf{76.5} $\pm$1.0 &	\textbf{97.3} $\pm$0.1 &	\textbf{99.4} $\pm$0.2&	 \textbf{99.4} $\pm$0.4&\textbf{63.0}$\pm$0.8\\
\bottomrule[1pt]
\end{tabular}
\label{Tab:AUC10}
\end{table*}

\subsection{Parameter Sensitivity Analysis}
\begin{figure*}[!htb]
\centering
\begin{tabular}{cc}
\includegraphics[scale=.43]{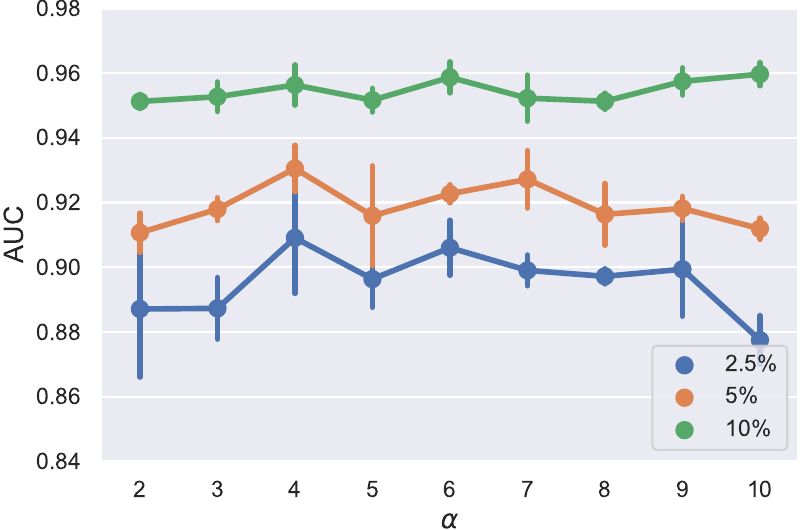} & \includegraphics[scale=.43]{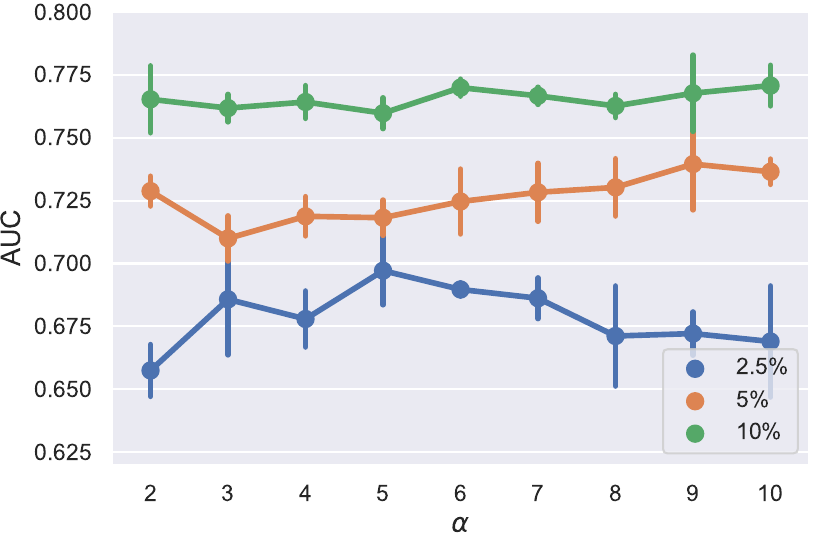}\\
(a) Cora & (b) Citeseer \\
\includegraphics[scale=.43]{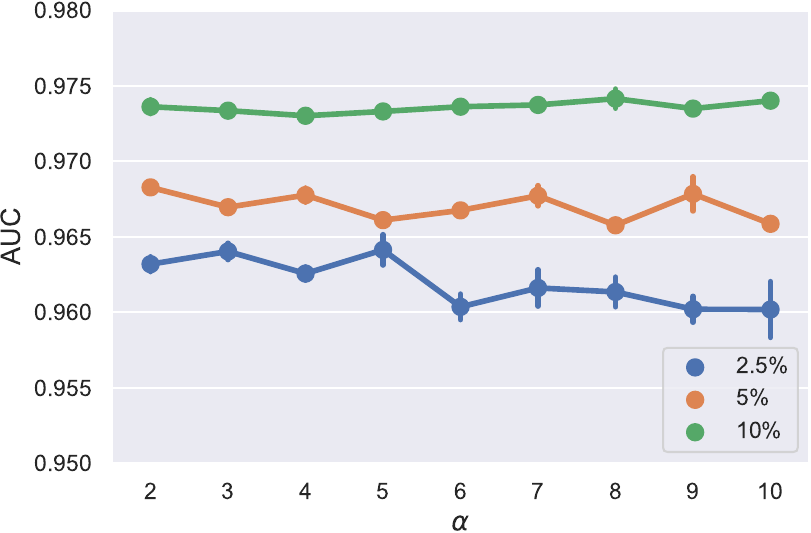} & \includegraphics[scale=.43]{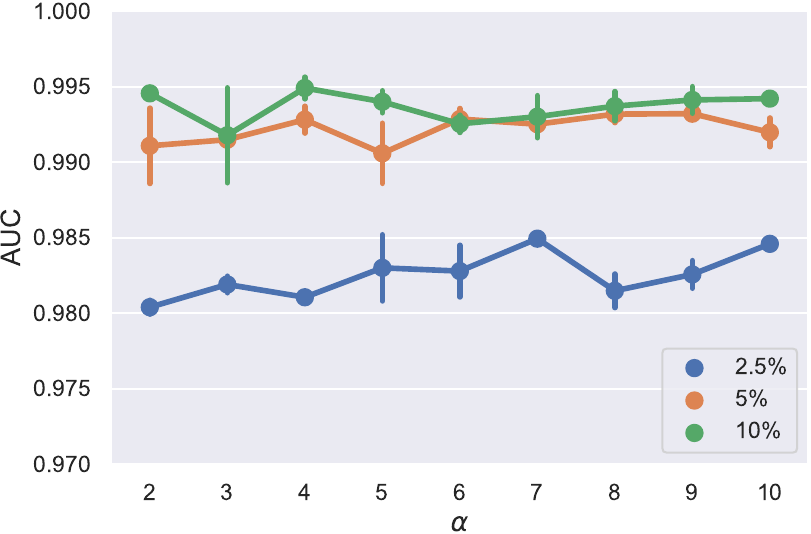}\\
(c) Pubmed & (d) Photo \\
\includegraphics[scale=.43]{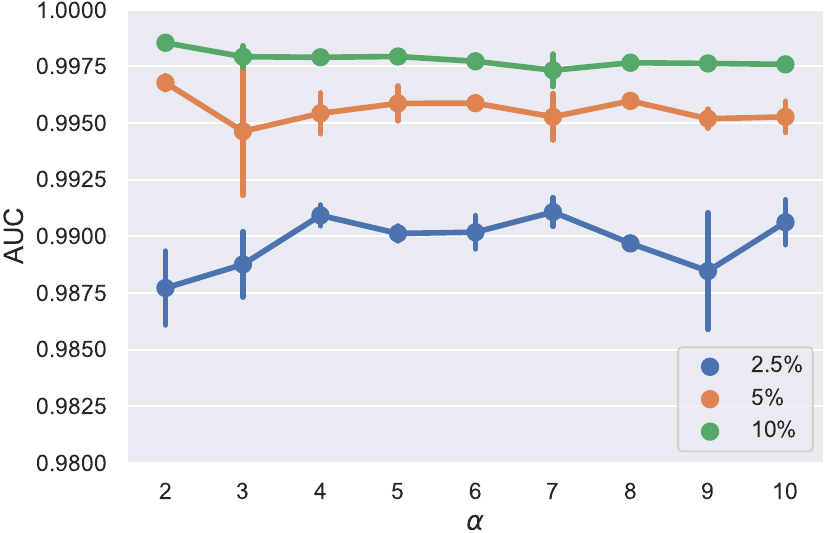} & \includegraphics[scale=.37]{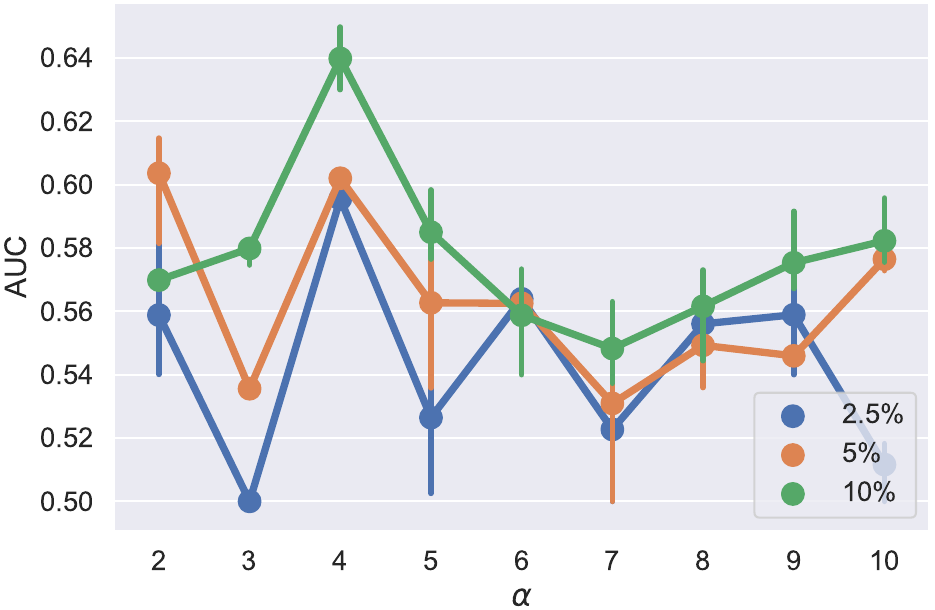}\\
(e) Computers & (f) ogbn-arxiv\\
\end{tabular}
\caption{Effect of weight hyperparameter $\alpha$ on anomaly detection performance (AUC).}
\label{Fig:alpha}
\end{figure*}

The weight hyperparameter $\alpha$ of the weighted cross-entropy and the regularization hyperparameter $\beta$ play an important role in the anomaly detection performance of the proposed GFCN framework. We conduct a sensitivity analysis to investigate how the performance of GFCN changes as we vary these two hyperparameters. In Figure~\ref{Fig:alpha}, we analyze the effect of the hyperparameter $\alpha$ by plotting the AUC results of GFCN vs. $\alpha$ using various label rates for all datasets, where $\alpha$ varies from 2 to 10. We can see that with a few exceptions, our model generally benefits from relatively larger values of the weight hyperparameter. For almost all datasets, our model achieves satisfactory performance with $\alpha=4$.

\begin{figure*}[!htb]
\centering
\begin{tabular}{cc}
\includegraphics[scale=.43]{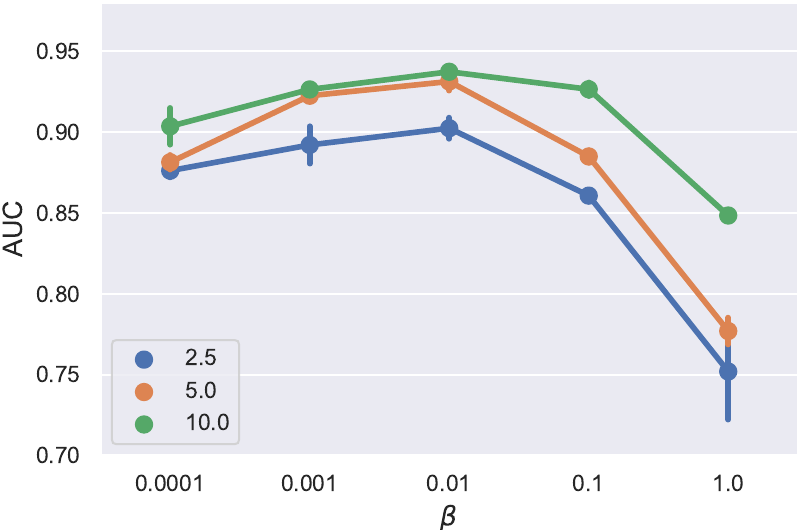} & \includegraphics[scale=.43]{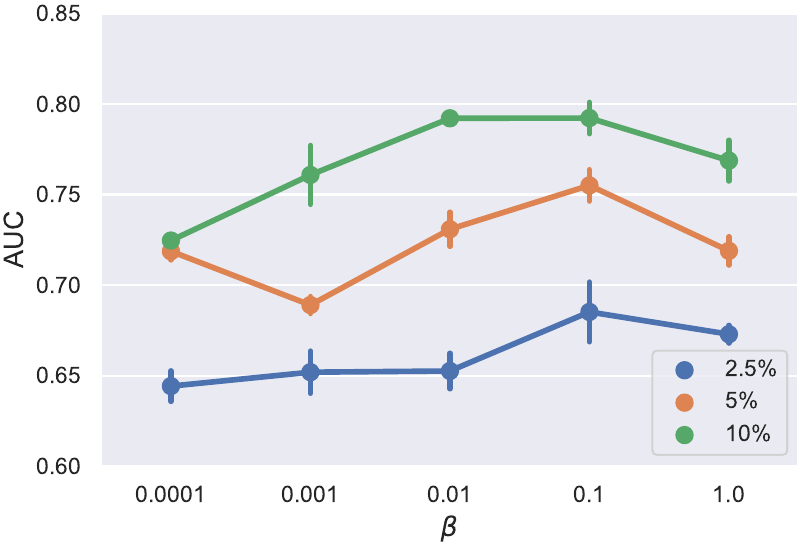}\\
(a) Cora & (b) Citeseer \\
\includegraphics[scale=.43]{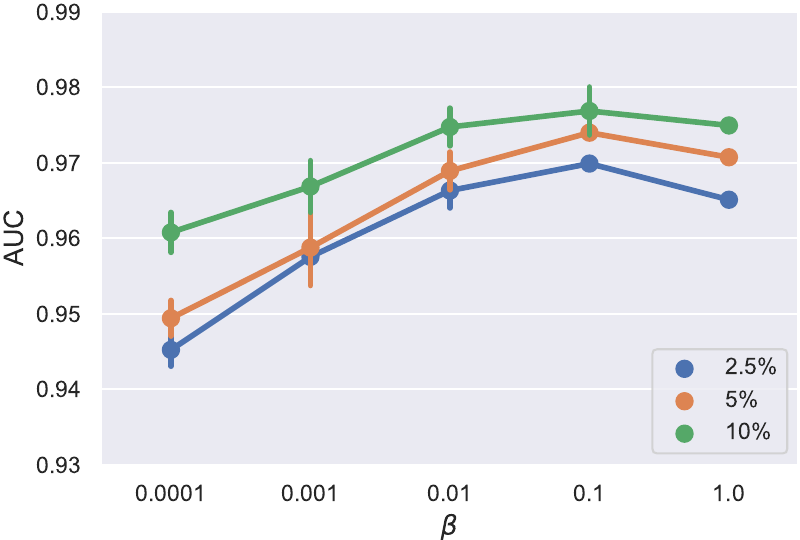} & \includegraphics[scale=.43]{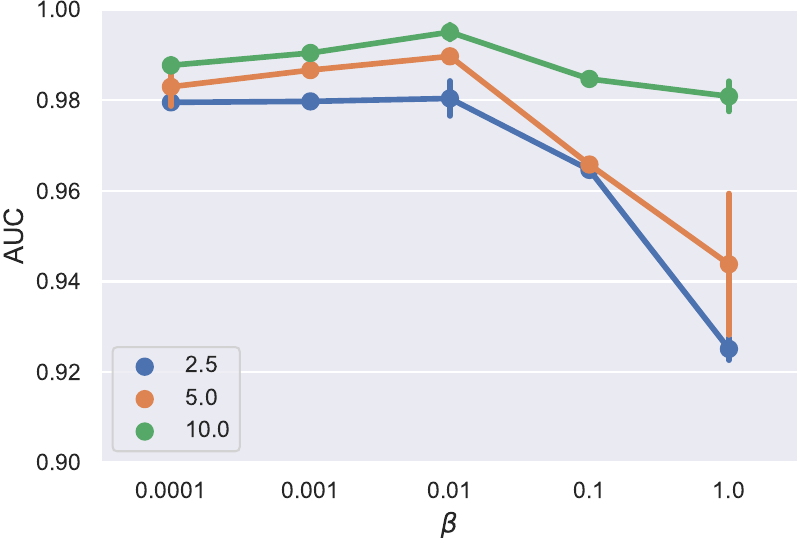}\\
(c) Pubmed & (d) Photo \\
\includegraphics[scale=.43]{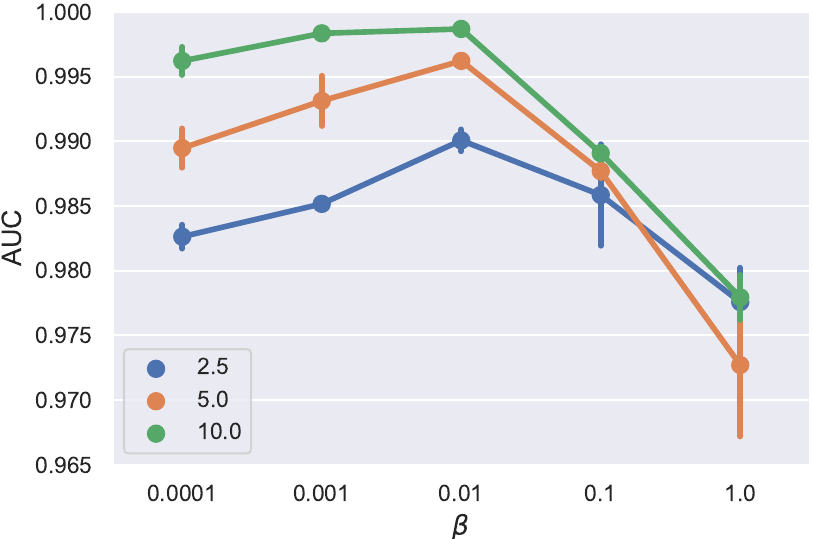} & \includegraphics[scale=.37]{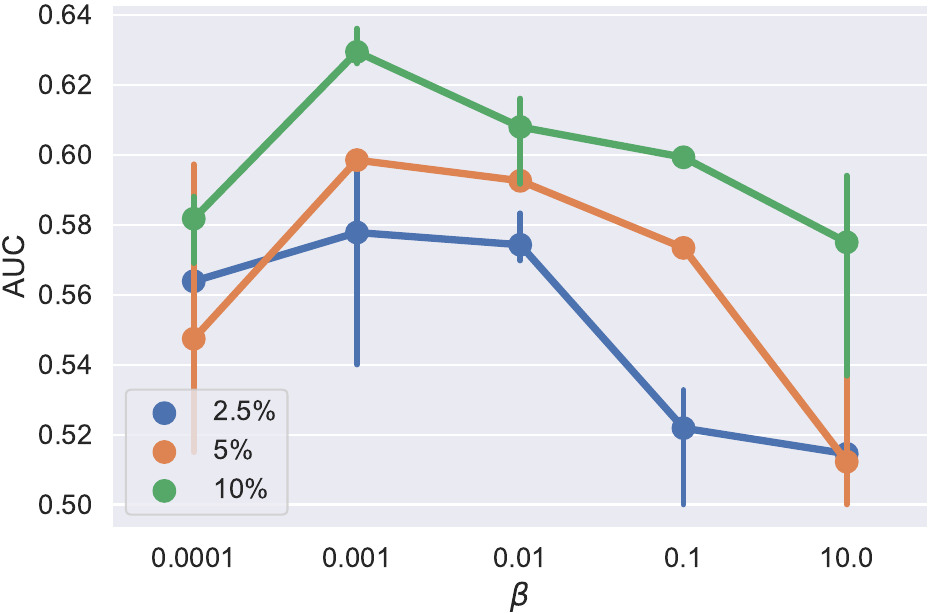}\\
(e) Computers & (f) ogbn-arxiv
\end{tabular}
\caption{Effect of regularization hyperparameter $\beta$ on anomaly detection performance (AUC).}
\label{Fig:beta}
\end{figure*}

In Figure~\ref{Fig:beta}, we plot the average AUCs, along with the standard error bars, of our GFCN model vs. $\beta$ using various label rates for all datasets, and by varying the value of $\beta$ from $10^{-4}$ to 1. Notice that the best performance is generally achieved when $\beta=0.01$, except in the cases of the Citeseer and Pubmed datasets, on which the best performance is obtained when $\beta=0.1$. In general, when the regularization hyperparameter increases, the performance improves rapidly at the very beginning, but then deteriorates after reaching the best setting due to overfitting. An interesting observation is that GFCN generally shows steady increase in performance with the regularization parameter, except on the Citeseer dataset when the label rate is 5\%, whereas the performance on the other datasets degrades after reaching a certain threshold.

\begin{figure*}[!htb]
\setlength{\tabcolsep}{1em}
\centering
\begin{tabular}{cc}
\includegraphics[scale=.55]{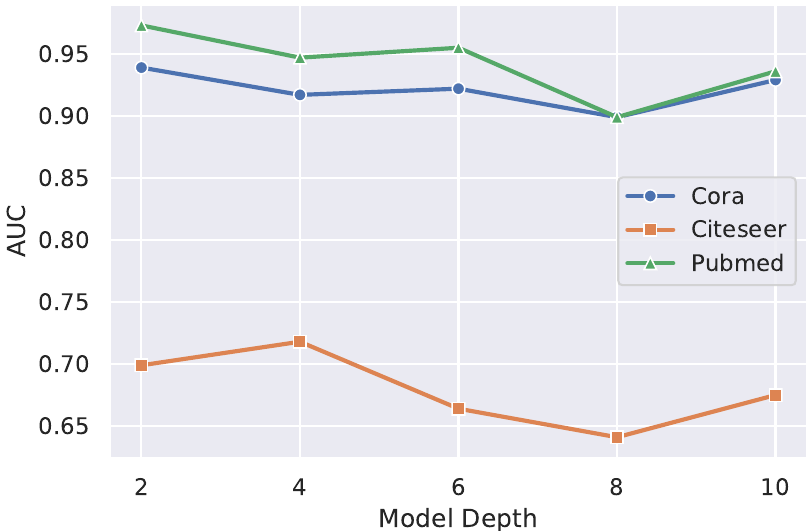} & \includegraphics[scale=.55]{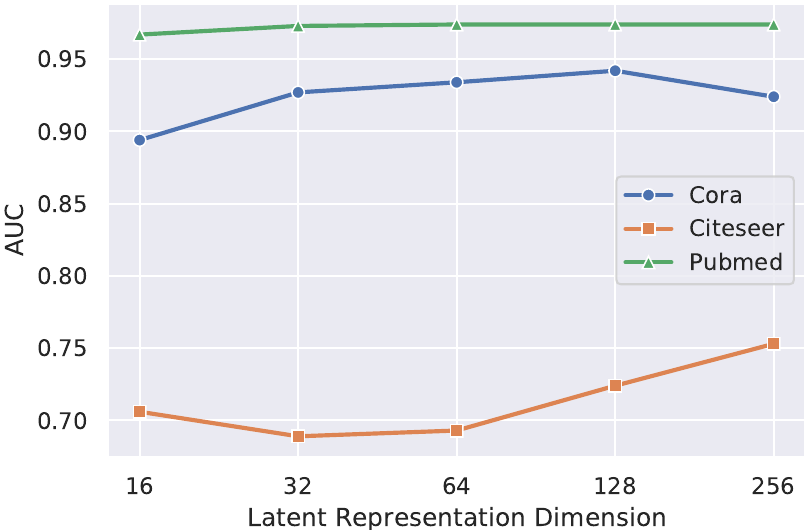}
\end{tabular}
\caption{Effects of number of layers (left) and latent representation dimension (right) on anomaly detection performance of our GFCN model using the Cora, Citeseer, Pubmed dataset when 10\% of instances are labeled. The AUC results are averaged over 10 runs.}
\label{Fig:depth-dim}
\end{figure*}

We also analyzed the effects of the number of layers and latent representation dimension on the performance of our model using the Cora, Citeseer, and Pubmed datasets (10\% of the instances as labeled), and the average AUC results are displayed in Figure~\ref{Fig:depth-dim}. As shown in Figure~\ref{Fig:depth-dim} (left), the performance of GFCN remains relatively stable as we increase the depth of the network from 2 to 10 layers. Figure~\ref{Fig:depth-dim} (right) shows the AUC results with the latent representation dimension varying from 10 to 256. As can be seen, our model typically benefits from larger latent dimensions, achieving a good performance with a latent representation dimension equal to 128 for all datasets.

\subsection{Visualization}
\begin{figure*}[!htb]
\setlength{\tabcolsep}{1em}
\centering
\begin{tabular}{cc}
\includegraphics[scale=.6]{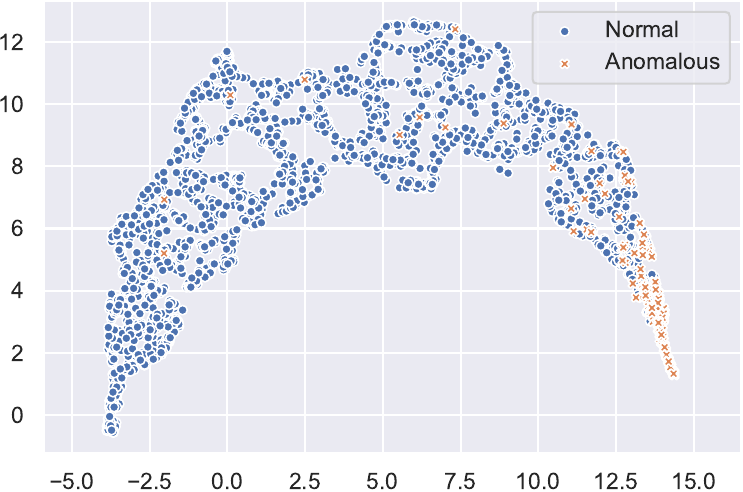} & \includegraphics[scale=.6]{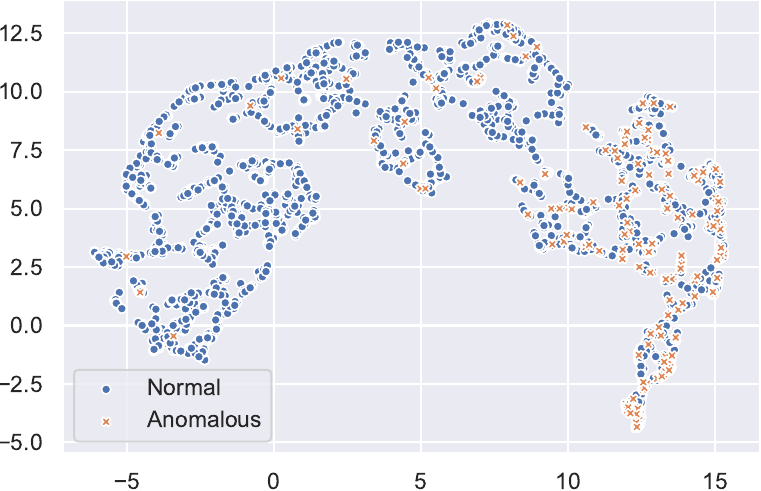}
\end{tabular}
\caption{UMAP embeddings of GFCN (left) and GCN (right) using the Cora dataset.}
\label{Fig:UMAP}
\end{figure*}

The feature embeddings learned by GFCN can be visualized using the Uniform Manifold Approximation and Projection (UMAP) algorithm~\cite{mcinnes2018umap}, which is a dimensionality reduction technique that is particularly well-suited for embedding high-dimensional data into a two- or three-dimensional space. Figure~\ref{Fig:UMAP} displays the UMAP embeddings of GFCN (left) and GCN (right) using the Cora dataset. As can be seen, the GFCN embeddings are more separable than the GCN ones. With GCN features, the normal and anomalous instances are not discriminated very well, while with GFCN features these data instances are discriminated much better, indicating that GFCN learns better node representations for anomaly detection tasks.

\subsection{Ablation Studies}
To validate the influence of our proposed components, we conduct additional experiments for ablation studies by removing components individually. The details and results of our experiments are shown in Table~\ref{Tab:Ablation study}, where we report both the average and standard deviation of AUCs over 10 runs on the Cora, Citeseer, and Pubmed datasets. As can be seen, the removal of each component leads to a deterioration in performance. By adding skip connections, an improvement in performance can be observed on all datasets. This indicates that reusing the initial node features in each layer and memorizing information across distant nodes yield improved results. Consistent with prior work, skip connections have been shown to mitigate the vanishing gradient problem in deep neural networks, as well as to improve the accuracy and convergence speed of the network. This has been demonstrated in a number of previous studies, including the original convolutional neural network with deep residual learning (ResNet)~\cite{He2016ResNet} and subsequent work on graph representation learning such as jumping knowledge network (JK-Net)~\cite{Keyulu:18}, residual graph convolutional network (ResGCN)~\cite{Li2019DeepGCN}, and graph convolutional networks with initial residual and identity mapping (GCNII)~\cite{Chen2020GCNII}. Also, Xu \textit{et al.}~\cite{Xu2021Optimization} examined the optimization dynamics of graph neural networks (GNNs) during their training process, and showed theoretically that  skip connections implicitly accelerate the training of GNNs.

As reported in Table~\ref{Tab:Ablation study}, the importance of the regularization term becomes apparent when removing it from the proposed loss function, resulting in a significant drop in AUC. Regularization is important to avoid model overfitting by imposing a penalty to learnable weights. Using regularization term in our loss function yields 9.2\%, 8.7\%, and 2.9\% performance gains over its counterpart model without regularization on the Cora, Citeseer, and Pubmed datasets, respectively.

On the other hand, it is worth pointing out that removing the skip connection and regularization components yields relatively high AUC standard deviations compared to the proposed GFCN model, indicating the robustness of our model.

\begin{table}[!htb]
\setlength{\tabcolsep}{2.3pt}
\caption{Test AUC (\%) averaged over 10 runs when 5\% of instances are labeled. We also report the standard deviation.}
\medskip
\centering
\begin{tabular}{lrrrrc}
\toprule[1pt]
\textbf{Method}&  Cora &  Citeseer & Pubmed\\
\midrule[.8pt]
Without skip connection & 91.2$\pm$2.7&64.6$\pm$3.8&94.2$\pm$0.2\\
Without regularization & 84.7$\pm$6.2&59.6$\pm$4.1&93.4$\pm$0.1\\
\midrule[.8pt]
\textbf{GFCN}&  \textbf{93.9}$\pm$2.3&	\textbf{68.3}$\pm$1.1 &	\textbf{96.3}$\pm$0.1\\
\bottomrule[1pt]
\end{tabular}
\label{Tab:Ablation study}
\end{table}

\section{Conclusion}
In this paper, we introduced a graph convolutional network with skip connection for semi-supervised anomaly detection on graph-structured data by learning effective node representations in an end-to-end fashion. The update rule of the proposed graph fairing convolutional network (GFCN) is theoretically motivated by implicit fairing and derived directly from the Jacobi iterative method. GFCN integrates skip connections between the initial feature matrix and each hidden layer. This allows our model to retain and reuse the original node features throughout the network, enabling better information propagation. We also showed that GFCN has the same time and memory complexity as the standard GCN, despite the inclusion of skip connections for improved node representations. Through extensive experiments, we demonstrated the competitive or superior performance of our model in comparison with the current state of the art on five benchmark datasets.While GFCN's intuitive design provides a solid theoretical foundation, it may face scalability challenges when dealing with very large graphs like many GCN-based methods. For future work, we plan to apply our approach to other downstream tasks on graph-structured data. We also intend to incorporate higher-order neighborhood information into the graph structure of the model, where nodes not only receive latent representations from their 1-hop neighbors, but also from multi-hop neighbors.

\section*{Acknowledgments}
This work was supported in part by the Discovery Grants program of Natural Sciences and Engineering Research Council of Canada.

\bibliographystyle{ieeetr}
\bibliography{references} 

\begin{thebibliography}{10}

\bibitem{Chandola:09}
V.~Chandola, A.~Banerjee, and V.~Kumar, ``Anomaly detection: A survey,'' {\em
  ACM Computing Surveys}, vol.~41, no.~3, pp.~1--58, 2009.

\bibitem{Guansong:20}
G.~Pang, C.~Shen, L.~Cao, and A.~van~den Hengel, ``Deep learning for anomaly
  detection: A review,'' {\em ACM Computing Surveys}, vol.~54, no.~2,
  pp.~1--38, 2021.

\bibitem{Doshi:21}
K.~Doshi and Y.~Yilmaz, ``Online anomaly detection in surveillance videos with
  asymptotic bound on false alarm rates,'' {\em Pattern Recognition}, vol.~114,
  2021.

\bibitem{Scholkopf:01}
B.~Sch\"{o}lkopf, J.~Platt, J.~Shawe-Taylor, A.~Smola, , and R.~Williamson,
  ``Estimating the support of a high-dimensional distribution,'' {\em Neural
  Computation}, vol.~13, no.~7, pp.~1443--1471, 2001.

\bibitem{tax2004support}
D.~M. Tax and R.~P. Duin, ``Support vector data description,'' {\em Machine
  learning}, vol.~54, no.~1, pp.~45--66, 2004.

\bibitem{Zhang2023Graph}
F.~Zhang, S.~Kan, D.~Zhang, Y.~Cen, L.~Zhang, and V.~Mladenovic, ``A graph
  model-based multiscale feature fitting method for unsupervised anomaly
  detection,'' {\em Pattern Recognition}, vol.~138, 2023.

\bibitem{Arias2023AIDA}
L.~A.~S. Arias, C.~W. Oosterlee, and P.~Cirillo, ``{AIDA}: Analytic isolation
  and distance-based anomaly detection algorithm,'' {\em Pattern Recognition},
  vol.~141, 2023.

\bibitem{wang2020deep}
R.~Wang, K.~Nie, T.~Wang, Y.~Yang, and B.~Long, ``Deep learning for anomaly
  detection,'' in {\em Proc. International Conference on Web Search and Data
  Mining}, pp.~894--896, 2020.

\bibitem{ruff2018deep}
L.~Ruff, R.~Vandermeulen, N.~Goernitz, L.~Deecke, S.~A. Siddiqui, A.~Binder,
  E.~M{\"u}ller, and M.~Kloft, ``Deep one-class classification,'' in {\em Proc.
  International Conference on Machine Learning}, pp.~4393--4402, 2018.

\bibitem{ruff2019deep}
L.~Ruff, R.~A. Vandermeulen, N.~G{\"o}rnitz, A.~Binder, E.~M{\"u}ller, K.-R.
  M{\"u}ller, and M.~Kloft, ``Deep semi-supervised anomaly detection,'' in {\em
  International Conference on Learning Representations}, 2019.

\bibitem{Defferrard:16}
M.~Defferrard, X.~Bresson, and P.~Vandergheynst, ``Convolutional neural
  networks on graphs with fast localized spectral filtering,'' in {\em Advances
  in Neural Information Processing Systems}, pp.~3844--3852, 2016.

\bibitem{Kipf:17}
T.~Kipf and M.~Welling, ``Semi supervised classification with graph
  convolutional networks,'' in {\em International Conference on Learning
  Representations}, pp.~1--14, 2017.

\bibitem{akoglu2015graph}
L.~Akoglu, H.~Tong, and D.~Koutra, ``Graph based anomaly detection and
  description: a survey,'' {\em Data Mining and Knowledge Discovery}, vol.~29,
  no.~3, pp.~626--688, 2015.

\bibitem{ding2019deep}
K.~Ding, J.~Li, R.~Bhanushali, and H.~Liu, ``Deep anomaly detection on
  attributed networks,'' in {\em Proc. SIAM International Conference on Data
  Mining}, pp.~594--602, 2019.

\bibitem{kumagai2020semi}
A.~Kumagai, T.~Iwata, and Y.~Fujiwara, ``Semi-supervised anomaly detection on
  attributed graphs,'' in {\em Proc. International Joint Conference on Neural
  Networks}, 2021.

\bibitem{Desbrun:99}
M.~Desbrun, M.~Meyer, P.~Schr\"{o}der, and A.~H. Barr, ``Implicit fairing of
  irregular meshes using diffusion and curvature flow,'' in {\em Proc.
  SIGGRAPH}, pp.~317--324, 1999.

\bibitem{Cevikalp2023Gap}
H.~Cevikalp, B.~Uzun, Y.~Salk, H.~Saribas, and O.~K\"{o}p\"{u}kl\"{u}, ``From
  anomaly detection to open set recognition: Bridging the gap,'' {\em Pattern
  Recognition}, vol.~138, 2023.

\bibitem{Donahue:17}
J.~Donahue, P.~Kr\"{a}henb\"{u}hl, and T.~Darrell, ``Adversarial feature
  learning,'' in {\em International Conference on Learning Representations},
  2017.

\bibitem{Schlegl:19}
T.~Schlegl, P.~Seeb\"{o}ck, S.~Waldstein, G.~Langs, and U.~Schmidt-Erfurth,
  ``{f-AnoGAN}: Fast unsupervised anomaly detection with generative adversarial
  networks,'' {\em Medical Image Analysis}, vol.~54, pp.~30--44, 2019.

\bibitem{DiMattia:19}
F.~D. Mattia, P.~Galeone, M.~D. Simoni, and E.~Ghelfi, ``A survey on {GANs} for
  anomaly detection,'' {\em arXiv preprint arXiv:1906.11632}, 2019.

\bibitem{Nalisnick:19}
E.~Nalisnick, A.~Matsukawa, Y.~W. Teh, D.~Gorur, and B.~Lakshminarayanan, ``Do
  deep generative models know what they don't know?,'' in {\em International
  Conference on Learning Representations}, 2019.

\bibitem{huang2022graph}
C.~Huang, M.~Li, F.~Cao, H.~Fujita, Z.~Li, and X.~Wu, ``Are graph convolutional
  networks with random weights feasible?,'' {\em IEEE Transactions on Pattern
  Analysis and Machine Intelligence}, 2022.

\bibitem{Keyulu:18}
K.~Xu, C.~Li, Y.~Tian, T.~Sonobe, K.~Kawarabayashi, and S.~Jegelka,
  ``Representation learning on graphs with jumping knowledge networks,'' in
  {\em Proc. International Conference on Machine Learning}, 2018.

\bibitem{gasteiger2018combining}
J.~Gasteiger, A.~Bojchevski, and S.~Günnemann, ``Combining neural networks
  with personalized pagerank for classification on graphs,'' in {\em Proc.
  International Conference on Learning Representations}, 2019.

\bibitem{Chen2020GCNII}
M.~Chen, Z.~Wei, Z.~Huang, B.~Ding, and Y.~Li, ``Simple and deep graph
  convolutional networks,'' in {\em Proc. International Conference on Machine
  Learning}, pp.~1725--1735, 2020.

\bibitem{Li2019DeepGCN}
G.~Li, M.~Muller, A.~Thabet, and B.~Ghanem, ``{DeepGCNs}: Can {GCNs} go as deep
  as {CNNs}?,'' in {\em Proc. IEEE International Conference on Computer
  Vision}, pp.~9267--9276, 2019.

\bibitem{Hamza2007IP}
Y.~Zhang and A.~B. Hamza, ``Vertex-based diffusion for {3-D} mesh denoising,''
  {\em IEEE Transactions on Image Processing}, vol.~16, pp.~1036--1045, 2007.

\bibitem{Emad2007GI}
E.~E. Abdallah, A.~B. Hamza, and P.~Bhattacharya, ``Spectral graph-theoretic
  approach to {3D} mesh watermarking,'' in {\em Proc. Graphics Interface},
  pp.~327--334, 2007.

\bibitem{Emad2009ISIVP}
E.~E. Abdallah, A.~B. Hamza, and P.~Bhattacharya, ``Watermarking {3D} models
  using spectral mesh compression,'' {\em Signal, Image and Video Processing},
  vol.~3, pp.~375--389, 2009.

\bibitem{Taubin:95}
G.~Taubin, ``A signal processing approach to fair surface design,'' in {\em
  Proc. SIGGRAPH}, pp.~351--358, 1995.

\bibitem{Taubin:96}
G.~Taubin, T.~Zhang, and G.~Golub, ``Optimal surface smoothing as filter
  design,'' in {\em Proc. European Conference on Computer Vision}, 1996.

\bibitem{Hammond:11}
D.~Hammond, P.~Vandergheynst, and R.~Gribonval, ``Wavelets on graphs via
  spectral graph theory,'' {\em Applied and Computational Harmonic Analysis},
  vol.~30, no.~2, pp.~129--150, 2011.

\bibitem{Levie:18}
R.~Levie, F.~Monti, X.~Bresson, and M.~M. Bronstein, ``{CayleyNets}: Graph
  convolutional neural networks with complex rational spectral filters,'' {\em
  IEEE Transactions on Signal Processing}, vol.~67, no.~1, pp.~97--109, 2018.

\bibitem{Bianchi:19}
F.~M. Bianchi, D.~Grattarola, C.~Alippi, and L.~Livi, ``Graph neural networks
  with convolutional {ARMA} filters,'' {\em IEEE Transactions on Pattern
  Analysis and Machine Intelligence}, vol.~44, no.~7, pp.~3496--3507, 2021.

\bibitem{Wijesinghe:19}
A.~Wijesinghe and Q.~Wang, ``{DFNets}: Spectral {CNNs} for graphs with
  feedback-looped filters,'' in {\em Advances in Neural Information Processing
  Systems}, 2019.

\bibitem{Kingma2015Adam}
D.~P. Kingma and J.~Ba, ``Adam: A method for stochastic optimization,'' in {\em
  International Conference on Learning Representations}, 2015.

\bibitem{Sen:08}
P.~Sen, G.~Namata, M.~Bilgic, L.~Getoor, and T.~{Eliassi-Rad}, ``Collective
  classification in network data,'' {\em AI Magazine}, vol.~29, no.~3,
  pp.~93--106, 2008.

\bibitem{shchur2018pitfalls}
O.~Shchur, M.~Mumme, A.~Bojchevski, and S.~G{\"u}nnemann, ``Pitfalls of graph
  neural network evaluation,'' in {\em Proc. Relational Representation Learning
  Workshop, NeurIPS}, 2018.

\bibitem{wu2018imverde}
J.~Wu, J.~He, and Y.~Liu, ``{ImVerde}: Vertex-diminished random walk for
  learning network representation from imbalanced data,'' in {\em Proc. IEEE
  International Conference on Big Data}, pp.~871--880, 2018.

\bibitem{feng2020graph}
W.~Feng, J.~Zhang, Y.~Dong, Y.~Han, H.~Luan, Q.~Xu, Q.~Yang, E.~Kharlamov, and
  J.~Tang, ``Graph random neural networks for semi-supervised learning on
  graphs,'' in {\em Advances in Neural Information Processing Systems}, 2020.

\bibitem{lee2022grafn}
J.~Lee, Y.~Oh, Y.~In, N.~Lee, D.~Hyun, and C.~Park, ``{GraFN}: Semi-supervised
  node classification on graph with few labels via non-parametric distribution
  assignment,'' in {\em Proc. SIGIR Conference on Research and Development in
  Information Retrieval}, 2022.

\bibitem{chen2021topology}
D.~Chen, Y.~Lin, G.~Zhao, X.~Ren, P.~Li, J.~Zhou, and X.~Sun,
  ``Topology-imbalance learning for semi-supervised node classification,'' in
  {\em Advances in Neural Information Processing Systems}, 2021.

\bibitem{Pickup16IJCV}
D.~Pickup, X.~Sun, P.~L. Rosin, R.~R. Martin, Z.~Cheng, Z.~Lian, M.~Aono, A.~B.
  Hamza, A.~Bronstein, M.~Bronstein, S.~Bu, U.~Castellani, S.~Cheng, V.~Garro,
  A.~Giachetti, A.~Godil, L.~Isaia, J.~Han, H.~Johan, L.~Lai, B.~Li, C.~Li,
  H.~Li, R.~Litman, X.~Liu, Z.~Liu, Y.~Lu, L.~Sun, G.~Tam, A.~Tatsuma, and
  J.~Ye, ``Shape retrieval of non-rigid {3D} human models,'' {\em International
  Journal of Computer Vision}, vol.~120, pp.~169--193, 2016.

\bibitem{Biasotti16VC}
S.~Biasotti, A.~Cerri, M.~Aono, A.~B. Hamza, V.~Garro, A.~Giachetti, D.~Giorgi,
  A.~Godil, C.~Li, C.~Sanada, M.~Spagnuolo, A.~Tatsuma, and
  S.~{Velasco-Forero}, ``Shape retrieval of non-rigid {3D} human models,'' {\em
  The Visual Computer}, vol.~32, pp.~217--241, 2016.

\bibitem{mcinnes2018umap}
L.~McInnes, J.~Healy, and J.~Melville, ``{UMAP}: Uniform manifold approximation
  and projection for dimension reduction,'' {\em The Journal of Open Source
  Software}, 2018.

\bibitem{He2016ResNet}
K.~He, X.~Zhang, S.~Ren, and J.~Sun, ``Deep residual learning for image
  recognition,'' in {\em Proc. IEEE Conference on Computer Vision and Pattern
  Recognition}, pp.~770--778, 2016.

\bibitem{Xu2021Optimization}
K.~Xu, M.~Zhang, S.~Jegelka, and K.~Kawaguchi, ``Optimization of graph neural
  networks: Implicit acceleration by skip connections and more depth,'' in {\em
  Proc. International Conference on Machine Learning}, 2021.

\end{thebibliography}

\end{document}